\documentclass[journal]{IEEEtran}
\usepackage{lineno}
\modulolinenumbers[5]

\makeatletter

\newcommand{\Rmnum}[1]{\expandafter\@slowromancap\romannumeral #1@}
\makeatother
\usepackage{xpatch}
\usepackage{bm}
\usepackage{array}
\usepackage{graphicx}
\usepackage{bigstrut}
\usepackage{amsmath,amssymb,amsthm}
\usepackage{algpseudocode}
\usepackage{algorithmicx}
\usepackage{algorithm}
\usepackage{booktabs}
\usepackage{color}
\usepackage{xcolor}
\usepackage{changepage}
\usepackage{xr}
\usepackage{xr-hyper}
\usepackage{graphicx}
\usepackage{subfigure}
\usepackage{caption}
\usepackage{makecell}
\usepackage{multirow}
\usepackage{float}
\DeclareCaptionLabelFormat{red}{\textcolor{red}{#1 #2}}
\usepackage{soul}
\usepackage[hidelinks]{hyperref}
\usepackage[numbers,sort&compress]{natbib}
\usepackage{bigstrut} 

 \newtheorem{definition}{Definition}
 
 \newtheorem{example}{Example}

\newtheoremstyle{nonbolditalic}
  {}
  {}
  {\normalfont}
  {}
  {\itshape}
  {.}
  {.5em}
  {\thmname{#1}\thmnumber{\textit{\enskip#2}}\thmnote{#3}}

\theoremstyle{nonbolditalic}
\newtheorem{proposition}{Proposition}

\AtBeginDocument{%
 \abovedisplayskip=5pt plus 4pt minus 2pt
 \abovedisplayshortskip=5pt plus 4pt minus 4pt
 \belowdisplayskip=5pt plus 4pt minus 2pt
 \belowdisplayshortskip=5pt plus 4pt minus 4pt
}

\ifCLASSINFOpdf

\else

\fi

\begin{document}
\captionsetup{font={footnotesize}}
\captionsetup[table]{labelformat=simple, labelsep=newline, textfont=sc, justification=centering}

\title{An Enhanced Differential Grouping Method \\ for Large-Scale Overlapping Problems

\author{Maojiang Tian,
        Mingke Chen,
        Wei Du,~\IEEEmembership{Member,~IEEE},
        Yang Tang,~\IEEEmembership{Fellow,~IEEE}, 
        and \\
        Yaochu Jin,~\IEEEmembership{Fellow,~IEEE}

\thanks{This work was supported by National Natural Science Foundation of China (Key Program: 62136003), National Natural Science Foundation of China (62173144, 62273149), the Program of Introducing Talents of Discipline to Universities (the 111 Project) under Grant B17017 and Shanghai AI Lab. (\emph{Corresponding author: Wei Du.})}

\thanks{M. Tian, M. Chen, W. Du and Y. Tang are with the Key Laboratory of Smart Manufacturing in Energy Chemical Process, Ministry of Education, East China University of Science and Technology, Shanghai 200237, China (e-mail: mjtian0618@gmail.com; mk.chen.ecust@gmail.com; duwei0203@gmail.com; yangtang@ecust.edu.cn).}

\thanks{Y. Jin is with the School of Engineering, Westlake University, Hangzhou 310030, China (e-mail: jinyaochu@westlake.edu.cn).}}}

\author{}

\markboth{}%
{Shell \MakeLowercase{\textit{et al.}}: Bare Demo of IEEEtran.cls for IEEE Journals}

\maketitle

\begin{abstract}
Large-scale overlapping problems are prevalent in practical engineering applications, and the optimization challenge is significantly amplified due to the existence of shared variables.
Decomposition-based cooperative coevolution (CC) algorithms have demonstrated promising performance in addressing large-scale overlapping problems.
However, current CC frameworks designed for overlapping problems rely on grouping methods for the identification of overlapping problem structures and the current grouping methods for large-scale overlapping problems fail to consider both accuracy and efficiency simultaneously.
In this article, we propose a two-stage enhanced grouping method for large-scale overlapping problems, called OEDG, which achieves accurate grouping while significantly reducing computational resource consumption.
In the first stage, OEDG employs a grouping method based on the finite differences principle to identify all subcomponents and shared variables.
In the second stage, we propose two grouping refinement methods, called subcomponent union detection (SUD) and subcomponent detection (SD), to enhance and refine the grouping results.
SUD examines the information of the subcomponents and shared variables obtained in the previous stage, and SD corrects inaccurate grouping results.
To better verify the performance of the proposed OEDG, we propose a series of novel benchmarks that consider various properties of large-scale overlapping problems, including the topology structure, overlapping degree, and separability.
Extensive experimental results demonstrate that OEDG is capable of accurately grouping different types of large-scale overlapping problems while consuming fewer computational resources.
Finally, we empirically verify that the proposed OEDG can effectively improve the optimization performance of diverse large-scale overlapping problems.
\end{abstract}


\IEEEpeerreviewmaketitle

\section{Introduction}
\IEEEPARstart{L}{arge-scale} global optimization (LSGO) problems are defined as optimization problems with thousands to billions of decision variables \cite{LSGO1}.
In the context of LSGO, a series of problem areas emerge \cite{LSGO2}.
Large-scale overlapping problems (LSOP), referring to optimization problems with shared variables between subcomponents, constitute a distinctive class of LSGO problems \cite{CEC2013}.
Such problems surface in numerous engineering application domains, including complex concurrent engineering \cite{concurrent}, automotive design \cite{car}, and water network optimization \cite{water}.

There are two types of evolutionary algorithms (EAs) for solving LSGO problems: non-decomposition-based methods and decomposition-based methods \cite{LSGO1, LSGO2}.
The non-decomposition-based approaches optimize all decision variables simultaneously.
To address the ``curse of dimensionality" caused by LSGO, recent non-decomposition-based methods aim at improving the exploration and exploitation of traditional EAs, e.g., competitive swarm optimizer (CSO) \cite{CSO}, gene targeting differential evolution (GTDE) \cite{GTDE}, and mixture model-based evolution strategy (MMES) \cite{MMES}.

The overlapping subcomponents in the overlapping problems establish connections through shared variables.
Exploring the relationship between subcomponents and implementing modular approaches to achieve rational problem grouping can enhance the optimization performance of such problems.
Extensive experiments have demonstrated that decomposition-based methods outperform non-decomposition-based methods for handling overlapping problems \cite{CBCCO, RDG3, DOV}.

The decomposition-based approaches, inspired by the concept of divide-and-conquer, utilize the cooperative coevolution (CC) \cite{CC1, CC2} framework to partition all variables into multiple groups before optimization.
CC optimizes each variable group in a round-robin fashion and employs context vectors to record the best solutions for each group until a termination condition is met.
In recent years, numerous enhanced CC frameworks have been proposed.
Contribution-based CC algorithms \cite{CBCC1, CBCC2, CBCC3, CCFR} achieve efficient allocation of computational resources according to the contributions of subgroups.
Difficulty and contribution-based CC (DCCC) \cite{DCCC} formulate an indicator to judge the difficulty of each sub-problem.
Two conformance policies are designed in \cite{DCCA} to guide the allocation of computational resources during the optimization process.

A series of decomposition methods have been proposed for large-scale global optimization problems.
In earlier studies, the decomposition method relied on static methods \cite{CC1, HALF, KS} and random decomposition methods \cite{CC4}, both of which were proved to be less accurate. 
Subsequently, a series of decomposition methods based on interaction detection significantly improved the accuracy of the decomposition process \cite{DG, IRRG, GSG}.

The most prevalent decomposition methods can be classified into three categories based on their principles of interaction detection: finite differences \cite{DG, RDG, EDG}, monotonicity checking \cite{CCVIL, FVIL, IRRG}, and the minimum points shift principle \cite{GSG, SVG}.
Among these, the differential grouping (DG) series methods, built upon the finite differences principle, are the most prevalent interaction detection methods \cite{MDG, ERDG, EADG, RDG3}.
In addition, based on the finite differences principle, some methods take advantage of the importance characteristics of the variables to conduct hybrid deep decomposition \cite{DCC, HDG, TSCC}.

However, most of the above decomposition methods fail to address overlapping problems.
This is because overlapping problems exhibit a complex interaction structure. 
All variables within each subcomponent directly interact with each other, and simultaneously, the presence of shared variables allows variables not belonging to the same subcomponent to interact indirectly.
Therefore, most decomposition methods classify all direct and indirect interaction variables into a single group, which complicates effective grouping and precise identification of the underlying structure for overlapping problems.

To leverage the structure of overlapping problems for efficient grouping, researchers have introduced a series of innovative DG series methods \cite{RDG3, ORDG, DG2}.
Traditional DG series methods continuously merge adjacent subcomponents when addressing overlapping problems, thereby resulting in an extensive size of the non-separable variable group.
To counteract this issue, recursive DG 3 (RDG3) \cite{RDG3}, an enhanced version of RDG, is particularly designed for overlapping problems. 
RDG3 applies a threshold to limit the size of subcomponents, effectively breaking down the overlapping problem into several smaller subproblems. 
However, the fixed thresholds in RDG3 are often determined by the researcher’s experience, which could lead to unreasonable grouping. 
Furthermore, RDG3 overlooks the identification of the subcomponents and overlapping variables in overlapping problems.

In contrast to using a fixed threshold for grouping overlapping problems, ORDG \cite{ORDG} aims to locate all shared variables and distribute them among adjacent subcomponents to preserve crucial interactions. 
ORDG first identifies a subcomponent and then employs the overlapping variables within that subcomponent to locate the second subcomponent.
This process repeats until all subcomponents and shared variables are discovered. 
However, ORDG exhibits both low accuracy and poor stability, primarily due to its high susceptibility to the topology type of overlapping problems and the selection of the initially detected variable.

In order to enhance the accuracy of grouping overlapping problems and identifying their underlying structure, some extended DG series methods \cite{GDG, GIAT, DG2} construct an interaction structure matrix to explore the interaction between all pairs of variables.
DG2 \cite{DG2} is representative of the grouping method based on the interaction structure matrix.
DG2 detects all variables in pairs following the finite differences principle to obtain the interaction structure matrix.
This matrix provides comprehensive information regarding the interactions among all variables, offering a complete representation of the structure of the overlapping problems and clarifying the relationships between variables.
The benefits of DG2 are maximized when addressing overlapping problems, due to its precise exploration of the problem structure.

A series of CC-based methods \cite{CBCCO, DCC, DOV}, aiming to solve overlapping problems, often utilize DG2 to gather information regarding the overlapping subcomponents and variables.
For example, the CBCCO method \cite{CBCCO}, which is an extended CC framework designed for overlapping problems, allocates shared variables to subcomponents with larger contributions to improve the optimization effect. 
DG2 can provide CBCCO with accurate information about subcomponents and shared variables, so DG2 is utilized as the grouping algorithm for CBCCO.
However, in order to produce the interaction structure matrix, DG2 examines all pairs of variables, making the computational complexity of DG2 up to $\mathcal{O}(N^2)$.
This is excessively high for the grouping stage, and it imposes a considerable burden on the entire optimization process.                   

In summary, three types of methods represented by RDG3, ORDG, and DG2 aim to group the large-scale overlapping problems.
Both RDG3 and ORDG have low grouping accuracy and fail to identify all subcomponents and shared variables to explore the underlying structure of the overlapping problems.
Moreover, the order of the variables chosen for detection significantly affects the grouping results for RDG3 and ORDG, thereby affecting the stability of the results.
In contrast, DG2 can precisely identify the structure of overlapping problems by detecting the interaction relationships among all pairs of variables.
Therefore, DG2 is applicable to any CC framework specifically designed for addressing overlapping problems.
However, DG2 uses excessive computational resources, which can negatively affect the subsequent optimization process.
In addition, it is worth mentioning that current research on overlapping issues is incomprehensive and insufficient.
Existing research on overlapping problems only involves the line topology structure and lacks consideration of the overlapping degree and separability of the problems. 
Therefore, it is in high demand to further analyze and explore the properties of overlapping problems.

To address the issues of low grouping accuracy and stability in RDG3 and ORDG, as well as the excessive computational resource consumption in DG2, this article proposes a two-stage enhanced differential grouping method called OEDG for large-scale overlapping problems. 
OEDG groups overlapping problems efficiently and accurately, and the resulting grouping can infer the underlying structure of overlapping issues.
OEDG comprises two stages: the first stage involves identifying all subcomponents in the overlapping problem by detecting interactions between the detected variable and the remaining variables. 
The second stage focuses on analyzing the grouping results from the first stage, identifying and correcting inappropriate grouping results to enhance the grouping accuracy.
With variable-to-set and set-to-set interaction detection, OEDG outputs decomposition results for overlapping problems that include all subcomponents and the shared variables in each subcomponent.
Furthermore, given the absence of comprehensive benchmarks for overlapping problems, this paper introduces a series of novel benchmarks to assess the performance of various methods in addressing overlapping problems.
The main contributions of this work are as follows:

\begin{itemize}

    \item
    We conduct a comprehensive study on large-scale overlapping problems by considering various topology structures of overlapping problems, including line, ring, and complex topologies. 
    Furthermore, we conduct a thorough investigation into the influence of overlapping degrees and the separability of these problems.

    \item 
    We propose a two-stage enhanced differential grouping method, named OEDG, for large-scale overlapping problems.
    By utilizing a two-stage cooperative mechanism, OEDG achieves, for the first time, the accurate and efficient decomposition and identification of the structure of overlapping problems with various topologies and overlapping degrees.
    The decomposition results of OEDG can be seamlessly integrated into cutting-edge CC-based optimization methods for overlapping problems.
    
    \item 
    In the first stage of OEDG, we design a novel and efficient grouping method to identify the subcomponents and shared variables in the overlapping problem, which ensures the efficiency of OEDG.
    To further enhance the stability of this grouping method, we introduce two grouping refinement methods in the second stage.
    These two approaches are designed to identify and rectify any incorrect grouping results, thereby improving the accuracy of OEDG.

    \item 
    We design a series of new overlapping benchmarks with diverse overlapping characteristics, which include different topology structures, overlapping degrees, and separability.
    OEDG is compared with the state-of-the-art methods on the proposed benchmarks.
    Extensive experiments show that OEDG can efficiently and accurately group problems and identify the structure of overlapping problems, leading to promising optimization results in solving such problems.
    
\end{itemize}

The remainder of the article proceeds as follows.
Section \ref{section Background} presents the relevant properties of the overlapping problems and the typical methods for problem decomposition and optimization.
Section \ref{section proposed} introduces the proposed OEDG in detail, including the problem grouping method and grouping refinement methods.
Section \ref{section Benchmark} presents a series of new overlapping benchmarks.
Next, the results of grouping and optimization experiments are presented in Section \ref{section Experimental}.
Finally, Section \ref{section Conclusion} concludes this article and outlines future research directions.

\section{Background}\label{section Background}

Efficient and accurate variable grouping plays a crucial role in solving large-scale optimization problems using the CC framework \cite{DG, CCVIL}. 

In this section, we introduce the concept of interaction and the finite differences principle for identifying interactions. 
Next, we will discuss three key properties associated with overlapping problems.
Additionally, we will discuss three typical grouping methods employed to address large-scale overlapping problems and consider their advantages and disadvantages.
Finally, we introduce several optimization methods designed for overlapping problems.

\subsection{Types of Interaction}\label{Section:Type of Interaction}
Interactions between variables can be classified as direct interaction or indirect interaction based on their interaction structure.
In overlapping problems, variables in the same subcomponent directly interact with each other, while variables in different subcomponents exhibit indirect interactions.

\begin{definition}\label{defn:Direct}
\emph{(Direct Interaction \cite{DG, XDG})}\\
In a differentiable objective function $f(\bm{x})$, decision variables $x_i$ and $x_j$ interact directly with each other if there exists a candidate solution $\bm{x^*}$, such that
\begin{equation}
\frac{\partial^{2} f(\bm{x^*})}{\partial x_{i} \partial x_{j}} \neq 0
\end{equation}
This is denoted by $x_i$ $\leftrightarrow $ $x_j$.
\end{definition}


\begin{definition}\label{defn:Indirect}
\emph{(Indirect Interaction \cite{XDG})}\\
In a differentiable objective function $f(\bm{x})$, decision variables $x_i$ and $x_j$ interact indirectly with each other if for every candidate solution $\bm{x^*}$, such that
\begin{equation}
\frac{\partial^{2} f(\bm{x^*})}{\partial x_{i} \partial x_{j}} = 0
\end{equation}
and a set of decision variables $(x_{k1},...,x_{kt}) \subset X$ exists, such that $x_i \leftrightarrow x_{k1} \leftrightarrow ... \leftrightarrow x_{kt} \leftrightarrow x_j$.
\end{definition}

\begin{example} \label{MSV}
A brief example of direct interaction and indirect interaction between two variables is $f(\bm{x})={\left(x_{1}-x_{2}\right)}^2 + {\left(x_{2}-x_{3}\right)}^2$.
$x_2$ interacts directly with both $x_1$ and $x_3$.
There is no direct interaction between $x_1$ and $x_3$. 
$x_1$ interacts indirectly with $x_3$ because both $x_1$ and $x_3$ interact directly with $x_2$.
Meanwhile, $f(\bm{x})$ is an overlapping problem with the shared variable $x_2$.
\end{example}

\subsection{Finite Differences Principle}\label{Section:Differential Grouping}
Currently, the most prevalent methods to identify the interactions between variables are DG series methods based on the finite differences principle.
DG \cite{DG} is the first grouping method that applies the finite differences principle.

\begin{proposition} \label{Pro1}
Let $f(\boldsymbol{x})$ be an additively separable function.
Two variables $x_p$ and $x_q$ in the function $f(\boldsymbol{x})$ interact with each other if the following condition holds:
\begin{eqnarray}
\begin{split}
\left.\Delta_{\delta, {x}_{p}}[f](\boldsymbol{x})\right|_{{x}_{p}=a, {x}_{q}=b_{1}} \neq\left.\Delta_{\delta, {x}_{p}}[f](\boldsymbol{x})\right|_{{x}_{p}=a, {x}_{q}=b_{2}}  \label{equation 3}
\end{split}
\end{eqnarray}
where
\begin{eqnarray}
\begin{split}
{\Delta_{\delta, {x}_{p}}[f](\boldsymbol{x})=f\left(\ldots, {x}_{p}+\delta, \ldots\right)-f\left(\ldots, {x}_{p}, \ldots\right)}
\end{split}
\end{eqnarray}
\end{proposition}

DG is based on variable-to-variable interaction detection, and RDG generalizes the finite differences principle to set-to-set interaction detection in order to improve efficiency.

\begin{proposition} \label{Pro1}
Let $\boldsymbol{X_{1}}$ and $\bm{X_{2}}$ represent two mutually exclusive groups of variables that are subsets of $\boldsymbol{X}$.
If there exist two unit vectors $\bm{u}_{1} \in U_{\boldsymbol{X_{1}}}$ and $\bm{u}_{2} \in U_{\bm{X_{2}}}$, and two real numbers $l_{1}, l_{2}>0$, along with a candidate solution $\bm{x}^{*}$ in the decision variable space, such that
\begin{eqnarray}
\begin{split}
\begin{aligned}
f\left(\bm{x}^{*}+l_{1} \bm{u}_{1}+l_{2} \bm{u}_{2}\right)- & f\left(\bm{x}^{*}{+}l_{2} \bm{u}_{2}\right)  \\
{\neq} & f\left(\bm{x}^{*}+l_{1} \bm{u}_{1}\right)-f\left(\bm{x}^{*}\right) \label{equation 5}
\end{aligned}
\end{split}
\end{eqnarray}
Then $\bm{X_{1}}$ interacts with $\bm{X_{2}}$.
\end{proposition}

RDG recursively examines set-to-set interaction relationships with a computational complexity of $\mathcal{O}(NlogN)$.
For brevity, the left side of Eq. (\ref{equation 5}) is denoted as $\Delta_1$, and the right side of Eq. (\ref{equation 5}) is denoted as $\Delta_2$.
Due to rounding errors associated with computer floating-point operations, Eq. (\ref{equation 5}) is transformed to $\left|\Delta_{1}-\Delta_{2}\right|>\epsilon$ for interaction detection.
Both DG and RDG employ a fixed threshold $\epsilon$, which often leads to poor grouping accuracy. 
However, RDG2 \cite{RDG2} enhances the grouping accuracy by utilizing an adaptive threshold that takes into account the fitness function value and the number of variables.

\subsection{Overlapping Problems}\label{Section: Overlapping Problems}


In this subsection, we introduce three properties of overlapping problems: topology structure, shared variable character, and overlapping degree.

\subsubsection{Topology Structure}
The topology structure represents the distribution information of multiple subgroups within an overlapping problem.
Specifically, it reflects which subcomponents each component overlaps with.
In this article, we mainly introduce three kinds of topology structures, including line topology, ring topology, and complex topology.

Almost all the existing overlapping benchmark problems are line topology problems \cite{CBCCO, DCCMAES, DOV}.
In the overlapping problems with line topology, all subcomponents, except for the first and last ones, interact exclusively with their two adjacent subcomponents on their sides, resulting in the formation of a line structure.

When the subcomponents on both ends of a line topology overlapping problem are connected, the line topology overlapping problem transforms into a ring topology overlapping problem.
In the ring topology overlapping problems, each subcomponent only interacts with two other subcomponents.

In addition to the above two topology types of overlapping problems, we collectively refer to the other topology types of overlapping problems as complex topology overlapping problems.
In these cases, a subcomponent might simultaneously interact with multiple other subcomponents. 
Additionally, some shared variables may be present in more than two subcomponents at the same time.
As a result, complex topology overlapping problems exhibit intricate structures, posing challenges in both decomposition and optimization.

\subsubsection{Shared Variable Character}

Overlapping problems can be categorized into two classes: the conforming overlapping problems and the conflicting overlapping problems \cite{CEC2013}.
The overlapping functions $f_{13}$ and $f_{14}$ in the CEC 2013 LSGO benchmark are Schwefel's functions with conforming overlapping subcomponents and conflicting overlapping subcomponents, respectively.

In conforming overlapping problems, the shared variables retain the same optimal value in the adjacent subcomponent functions.
Consequently, when optimizing a certain subcomponent in the conforming overlapping problems, the optimized shared variables contribute to the optimization of the other subcomponents.
However, in conflicting overlapping problems, the optimal values of shared variables differ across subcomponent functions.
Conflicting shared variables may lead to the optimization of one subcomponent negatively affecting the other overlapping subcomponents.
Understanding the underlying structure of conforming and conflicting overlapping problems, along with properly grouping subcomponents and shared variables, can considerably enhance the optimization process \cite{DOV, CBCCO}.

\subsubsection{Overlapping Degree}

In this section, we first provide a definition of the overlapping degree.
Then, we will discuss the methods for setting the line topology overlapping problems with different overlapping degrees in existing research.

\begin{definition}\label{defn:OP}
\emph{(Overlapping Degree)}\\
In an $n$-dimensional overlapping function $f(\bm{x})$ with $k$ overlapping variables, the overlapping degree (OD) of the problem is defined as:
\begin{equation}
OD=\frac{k}{n}
\end{equation}
\end{definition}

The overlapping degree represents the proportion of the shared variables among all variables in the overlapping problems.
CEC 2013 LSGO [3] sets an overlapping size parameter, denoted as $m$, to control the number of shared variables in two adjacent subcomponents.
Calculating the overlapping degree relies on accurate identification of the problem structure.
For example, in an $n$-dimensional line topology overlapping problem containing $j$ subcomponents with an overlapping size $m$, there are ${(j-1)\times{m}}$ overlapping variables, so the overlapping degree is given by $\frac{(j-1)\times{m}}{n}$.

\subsection{Grouping Methods for Overlapping Problems}
In this subsection, we introduce three methods (RDG3, ORDG, DG2) based on the finite differences principle for overlapping problems.
We will also analyze their grouping process, advantages, and disadvantages, providing corresponding legends for clarity.

\subsubsection{RDG3}
Based on the finite differences principle, RDG can efficiently group non-separable variables using the set-to-set interaction detection method.
However, overlapping problems often involve a single non-separable variable group of significant size, posing a challenge for RDG to divide them into smaller subcomponents.
RDG3 makes some improvements in the process of the RDG algorithm, aiming at solving the overlapping problems efficiently.

RDG3 sets a threshold $\epsilon_{n}$ to control the size of the group of non-separable variables generated during the grouping process.
If the current variable group size is larger than the threshold $\epsilon_{n}$, RDG3 stops the subsequent interaction detection process and selects a new variable for grouping. 
By setting the threshold value, the overlapping problem can be divided into several subcomponents with smaller sizes.
The choice of the threshold $\epsilon_{n}$ has a crucial impact on the variable grouping results of RDG3.
We present an example in Section S.I.A of the Supplementary Material to illustrate the impact of the threshold on the RDG3 grouping process.

\subsubsection{ORDG}

The primary strategy of ORDG is to identify subcomponents and distribute shared variables among adjacent subcomponents for grouping.

ORDG first identifies a specific subcomponent along with the shared variables within it. 
Then, it leverages these shared variables to locate the subsequent subcomponent.
These two processes are repeated until all variables are grouped.
Finally, ORDG attempts to obtain each subcomponent in the overlapping problem, and each subcomponent contains all its shared variables.
To explain the ORDG process, we utilize the example in Section S.I.B of the Supplementary Material.

The choice of the initial detected variable has a great impact on the grouping accuracy of ORDG.
This drawback of ORDG grouping is also evident in Section S.I.B of the Supplementary Material.
ORDG is exclusively designed for addressing line overlapping problems and may not be effective in resolving overlapping problems with other topology structures.
As a result, ORDG exhibits poor accuracy and weak stability. 

\subsubsection{DG2}
To construct the interaction structure matrix, DG2 detects all pairwise variable interactions, giving it the advantage of accurately determining the interaction structure of overlapping problems.

The first part of the DG2 process involves calculating the difference $\left|{\Delta_1}-{\Delta_2}\right|$ for all pairs of variables as per Eq. (\ref{equation 3}) and recording this in the raw interaction structure matrix. 
The raw interaction structure matrix is then converted into an interaction structure matrix based on the threshold between pairs of variables.
According to the interaction structure matrix, we can discern the interaction relationship between all variables in the problem.
We show an example of an interaction structure matrix generated by DG2 for an overlapping problem in Section S.I.C of the Supplementary Material.

\subsection{Optimization Methods for Overlapping Problems}
Based on the above decomposition methods, a variety of CC-based optimization methods have been proposed.

RDG3 randomly assigns shared variables to specific subcomponents.
This leads to the formation of multiple non-overlapping groups of non-separable variables.
These groups of variables are directly optimized using the CC framework.
CC-RDG3 efficiently decomposes overlapping problems, enhancing optimization performance by employing a divide-and-conquer approach.
CC-RDG3 won first place in the CEC 2019 LSGO competition \cite{RDG3}.

The optimization process of ORDG takes into account the challenge of allocating shared variables.
During the optimization of each subcomponent, ORDG assigns overlapping variables to their respective subcomponents, resulting in the generation of multiple candidates. 
Subsequently, ORDG selects the individual with the best performance as the allocation scheme of shared variables.
The context vectors are updated according to the solution obtained in this allocation scheme.

The optimization methods based on the DG2 method, such as DOV \cite{DOV}, DCC \cite{DCC}, and CBCCO \cite{CBCCO}, can obtain more accurate structure information of overlapping problems.

DOV utilizes the interaction structure matrix obtained from DG2 to break the overlapping problem into several overlapping subcomponents.
These subcomponents are subsequently merged to reduce the number of shared variables between subcomponents.
Then, DOV proposes three strategies sharing information, mean value, and random selection to decide the value of the shared variables in the CC framework.

DCC first collects the contribution information of each dimension through the history-based overall fitness value obtained during the random grouping optimization process.
Then, the subcomponents are dynamically generated based on the interaction information and contribution information.
DCC groups variables with large contributions into one group and allocates more computational resources to them.
Additionally, a stage-by-stage parameter adaptation strategy is proposed to update the parameters of the dynamically changing subcomponents at a high frequency.

After using DG2 to identify all subcomponents and shared variables, CBCCO introduces a contribution-based method, which assigns shared variables to subcomponents with more significant contributions. 
CBCCO implements a novel contribution-based CC framework that utilizes an efficient reward scheme to reward the important subcomponents during the CC process. 
This framework efficiently allocates computational resources and maintains a high level of cooperation frequency among the optimizers.

Extensive experiments demonstrate that CBCCO outperforms other algorithms when optimizing large-scale overlapping problems using subcomponent and shared variable information \cite{CBCCO}.
Therefore, in this article, we select CBCCO as the optimization framework for the proposed decomposition method OEDG.

\section{The Proposed OEDG}\label{section proposed}
To overcome the limitations of previous grouping methods for large-scale overlapping problems, we propose the overlapping enhanced differential grouping (OEDG) method.
OEDG accurately groups overlapping problems, determining the grouping results of subcomponents and overlapping variables within each subcomponent.
Moreover, it allows for the deduction of the underlying structure of overlapping problems. 
Additionally, OEDG relies on variable-to-set and set-to-set interaction detection, which consumes fewer computational resources.

In this section, we first introduce the general framework of OEDG. 
Next, we present an effective grouping method for large-scale overlapping problems.
We then analyze the factors that may impact the accuracy of the grouping results and propose two collaborative methods to enhance algorithm stability.
Finally, we conduct the time complexity analysis of OEDG.

\subsection{The Framework of OEDG}
OEDG consists of two crucial stages: the problem grouping stage and the grouping refinement stage. 
Their cooperative approach is essential for the overall effectiveness of the algorithm. 
The first stage focuses on optimizing efficiency, while the second stage refines the accuracy of grouping results. 
OEDG's objective is to identify subcomponents and shared variables, allowing for a more tailored application in CC frameworks designed for overlapping problems. 
The pseudocode for OEDG can be found in Algorithm \ref{algorithm_OEDG}.

The OEDG process begins with the problem grouping stage, which outputs the subcomponents and shared variables within each formed subcomponent (Lines 2-12).
It begins by initializing the set of overlapping variable groups $OV$, the set of subcomponents $N$, and the ungrouped variable group $V_1$ (Line 2).
Subsequently, the algorithm calculates the fitness function value corresponding to the lower bound of $\emph{\textbf{x}}$ to reduce computational resource consumption during subsequent detection (Line 3). 
It then proceeds to find all subcomponents and shared variables from the overlapping problem, iterating until all variables have been grouped (Lines 4-12).

However, the grouping results generated in the first stage may not be completely accurate, as some subcomponents may be a union of several actual subcomponents.
The second stage applies a grouping refinement mechanism to optimize the grouping results and enhance the algorithm stability (Lines 14-21).
The subcomponent union detection (SUD) method sequentially examines whether all formed subcomponents in $N$ can be further decomposed, while subcomponent detection (SD) breaks down certain subcomponents into smaller and more accurate ones.
Finally, OEDG outputs accurate subcomponents and shared variables information for overlapping problem.
The specific details of the two stages are described in the following subsections.

\begin{algorithm}[!h]
\small
\caption{OEDG}
\label{algorithm_OEDG}
\begin{algorithmic}[1]
\Require{\emph{f}, $V$ (all variables), $\emph{\textbf{ub}}$, $\emph{\textbf{lb}}$}
\Ensure{$N$ (a set of subcomponents), $OV$ (a set of shared variable groups)}
\State /****** Problem Grouping Stage ******/
\State $OV \gets \{\}$, $N \gets \{\}$, $V_1$ $\gets$ $V$
\State $\emph{\textbf{x}}_{l,l} \gets \emph{\textbf{lb}}$, $\emph{{f}}_{l,l} \gets \emph{{f}}(\emph{\textbf{x}}_{l,l})$ 
\While {$V_1$ is not empty}
\State $X_1$ $\gets$ Pick a variable $x_i$ from $V_1$ at random
\State $X_1$ $\gets$ INTERACT ($X_1$, $V$, $\emph{\textbf{x}}_{l,l}$, $\emph{{f}}_{l,l}$, $\emph{\textbf{ub}}$, $\emph{\textbf{lb}}$)
\State $N$ $\gets$ $N$ $\bigcup$ $X_1$
\State $V_1$ $\gets$ $V_1$ $/$ $X_1$
\State $X_2$ $\gets$ $V$ $/$ $X_1$
\State $X_{OV}$ $\gets$ INTERACT-OV ($X_1$, $X_2$, $\emph{\textbf{x}}_{l,l}$, $\emph{{f}}_{l,l}$, $\emph{\textbf{ub}}$, $\emph{\textbf{lb}}$)
\State $OV$ $\gets$ $OV$ $\bigcup$ $X_{OV}$
\EndWhile  
\State /****** Grouping Refinement Stage ******/
\State $H$ $\gets$  Variables that occur twice in $OV$
\State $k$ $\gets$ length$(N)$
\For {$i \gets 1$ to $k$}
\While{SUD ($i$, $k$, $OV$)}
\State ($N$, $OV$) $\gets$ SD ($i$, $OV$, $N$, $H$, $V$, $\emph{\textbf{x}}_{l,l}$, $\emph{{f}}_{l,l}$, $\emph{\textbf{ub}}$, $\emph{\textbf{lb}}$)
\State $k$ $\gets$ $k$ + 1
\EndWhile
\EndFor
\State\Return $N$, $OV$
\end{algorithmic} 
\end{algorithm}

\begin{figure*}[htbp]
\centerline{\includegraphics[width=0.9\linewidth]{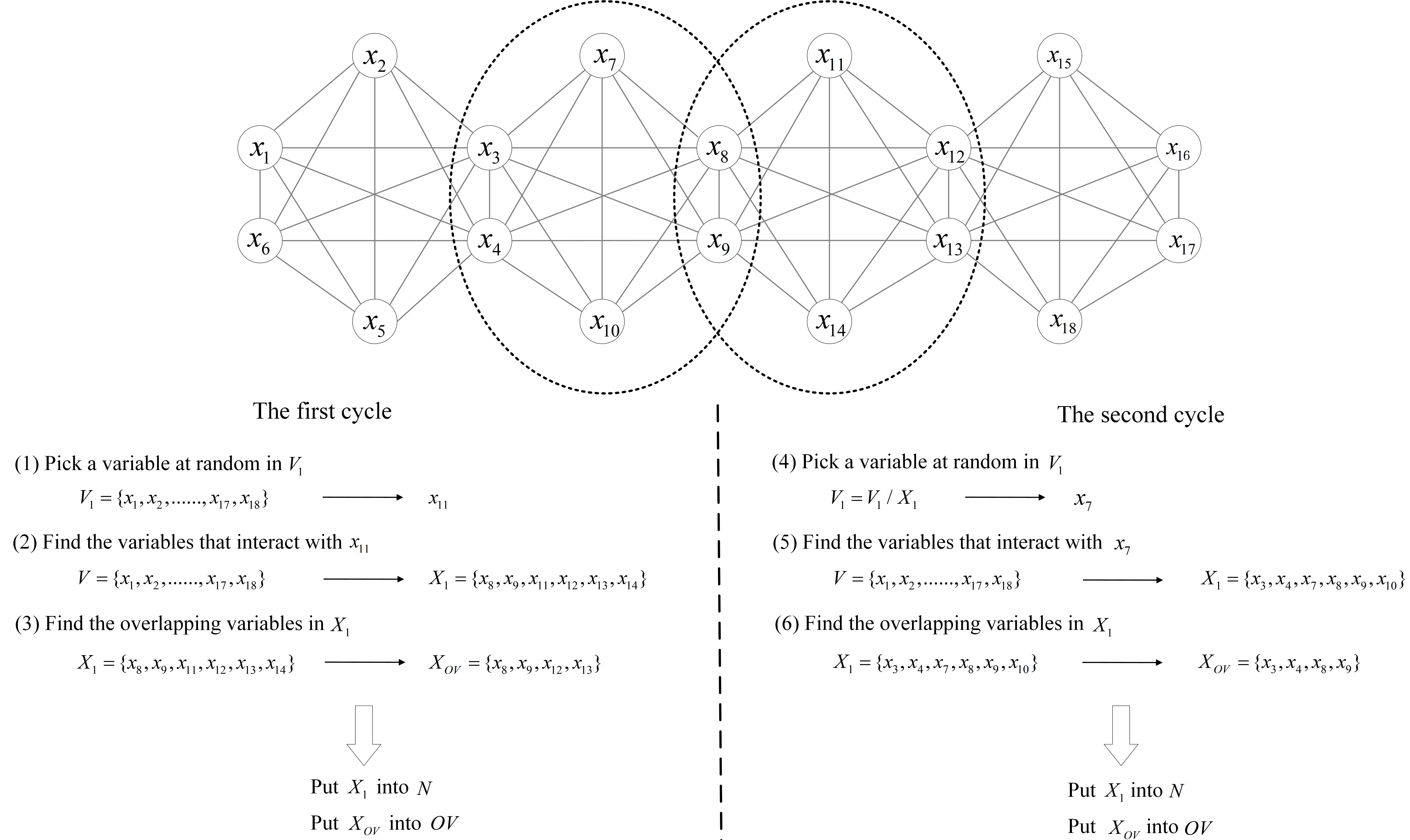}}
\caption{An example to illustrate the grouping process of OEDG, with the legend depicting the initial two cycles of grouping.}
\label{pic4}
\vspace{-0.35cm}
\end{figure*}

\subsection{Stage I: Problem Grouping}

The objective of the first stage is to identify each subcomponent in the overlapping problem and ascertain the shared variables present in each subcomponent.
OEDG executes multiple cycles in the first stage, and each cycle outputs a subcomponent and its shared variables.
The pseudocode of the problem grouping phase is shown in Algorithm \ref{algorithm_OEDG}.
In each cycle of the problem grouping process, the subcomponents and shared variables are detected based on two interaction detection methods.
The details of these interaction detection methods are shown in Section S.II of the Supplementary Material.

In the overlapping problem grouping stage, OEDG randomly selects a variable $x_i$ from the group of ungrouped variables as the detected variable (Line 5).
The variables in $V$ that directly interact with the detected variable $x_i$ are then identified, as shown in Algorithm S.1 in the Supplementary Material.
These variables are used to construct a subcomponent, which is denoted as $X_1$ (Line 6).
We record the grouping result $X_1$ into $N$ and remove the variables in $X_1$ from the ungrouped variable group $V_1$ (Lines 7-8).
We then identify the variables that interact between the variable group $X_1$ and the other variables, referred to as the overlapping variables $X_{OV}$ (Lines 9-11).
The method for identifying shared variables from each subcomponent is shown in Algorithm S.2 in the Supplementary Material.
Each grouping cycle ends when a subcomponent and a shared variable group are identified, and the next cycle begins by randomly selecting a variable to be detected from the remaining undetected variables $V_1$.
This grouping process is repeated until all variables in $V_1$ have been grouped into their respective subcomponents.
We finally obtain a set of subcomponents and a set of shared variable groups in each subcomponent, with the groups of variables in the two sets corresponding in order.

We present a diagram, illustrated in Fig. \ref{pic4}, to visually depict the complete process of the first stage.

In the given example of the overlapping problem illustrated in Fig. \ref{pic4}, a total of 18 variables can be divided into four subcomponents.
During each cycle, a detected variable $x_i$ is randomly selected from the remaining undetected variables.
Let's assume that in this case, the first detected variable is $x_{11}$.
The variables $\{x_8, x_9, x_{12}, x_{13}, x_{14} \}$ that interact with $x_{11}$ are identified using a recursive differential grouping method (Algorithm S.1) to form the first subcomponent $X_1$.
After identifying the first subcomponent, we proceed to determine the shared variables within this subcomponent.
We detect the interaction between the group of variables $X_1$ and the remaining variables ($V / X_1$) using Algorithm S.2.
Then, we identify the overlapping variables $X_{OV}=\{x_8, x_9, x_{12}, x_{13} \}$ in $X_1$.
In the second cycle, we assume that the selected variable to be detected is $x_7$.
After two interaction detections, the subcomponent $\{x_3, x_4, x_7, x_8, x_{9}, x_{10}\}$ and shared variables $\{x_3, x_4, x_8, x_{9}\}$ are sequentially identified.
We can observe that the two shared variable groups obtained in the above two rounds of grouping involve the same variables $x_8$ and $x_9$.
This suggests that the two subcomponents obtained from the grouping procedure overlap.
The grouping process ends when there are no remaining variables in $V_1$.
Finally, we obtain a set of subcomponents and a set of shared variable groups in each subcomponent.

\subsection{Stage II: Grouping Refinement} 

\begin{figure}[hbt!]
        \centering
        \subfigure[Select $x_5$ as the detected variable]{
            \begin{minipage}[b]{0.5\textwidth}
            \centering
            \includegraphics[width=0.825\textwidth]{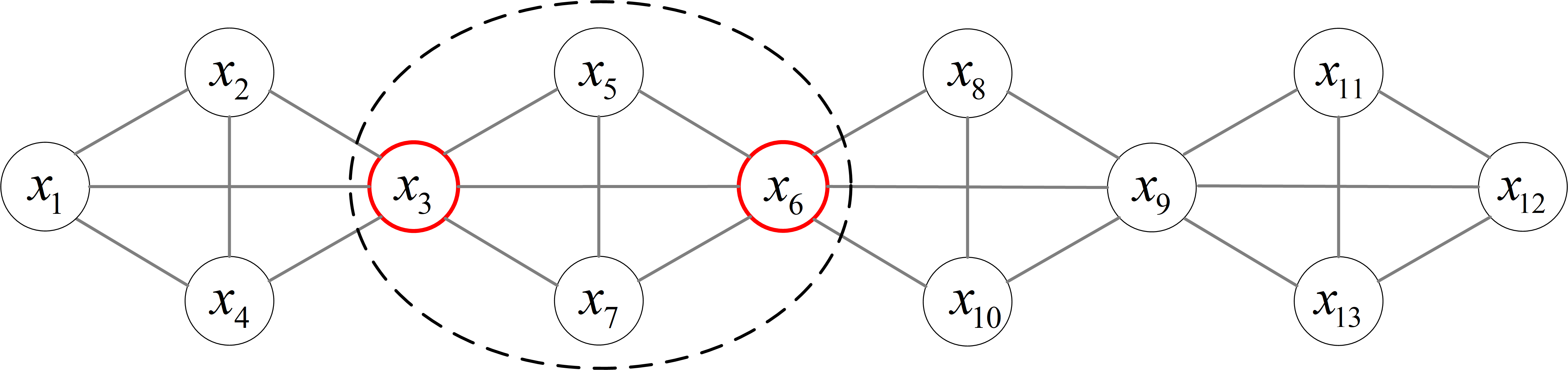}
            \end{minipage}
            }
            \hfill
        \subfigure[Select $x_6$ as the detected variable]{
            \begin{minipage}[b]{0.5\textwidth}
            \centering
            \includegraphics[width=0.825\textwidth]{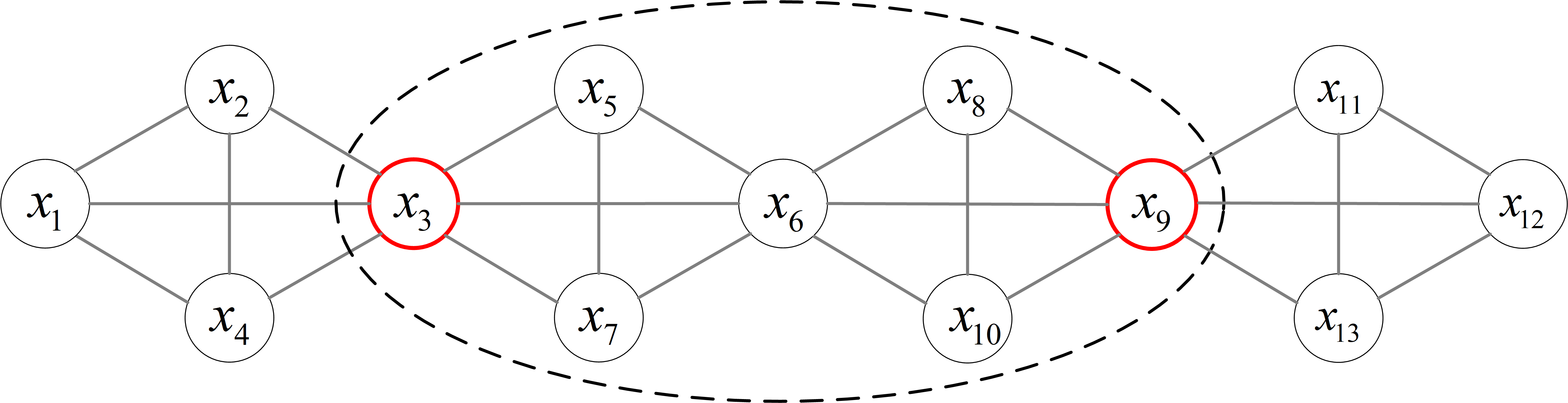}
            \end{minipage}
            }
        \caption{An example illustrating how the selection of shared variables can impact the grouping results.}
        \label{fig:refine}
          \vspace{-0.25cm}
\end{figure}

The second stage aims to refine the grouping results and improve the algorithm's stability.
Since $x_i$ is randomly selected from the ungrouped variable group $V_1$, an inappropriate selection may result in inaccuracies in some formed subcomponents in $N$.
This situation may necessitate further decomposition of certain subcomponents.
To illustrate this problem, we provide an example depicted in Fig. 
\ref{fig:refine}.

For the overlapping problem in Fig. \ref{fig:refine}, if the selected detected variable is $x_5$, then the formed subcomponent $X_1$ will be $\{x_3, x_5, x_6, x_{7}\}$.
But if we choose $x_6$ as the detected variable, the grouping result of $X_1$ is $\{x_3, x_5, x_6,...,x_{10} \}$.
In Fig. \ref{fig:refine}(a), all variables in the subcomponent $X_1$ are directly interacting, and the subcomponent cannot be further decomposed.
However, the subgroup formed in Fig. \ref{fig:refine}(b) is the union of two real subcomponents and can be further decomposed.
The reason for this is that the shared variable interacts with multiple adjacent subcomponents, leading to a detected variable merging multiple subcomponents, which affects the accuracy and stability of OEDG in identifying overlapping problem structures.
ORDG and RDG3 similarly suffer from the problem that the selection of the detected variables affects the grouping accuracy.

\begin{algorithm}[!h]
\small
\captionsetup{labelfont={color=black}}
\caption{SUD (Subcomponent Union Detection)}
\label{algorithm_SUD}
\begin{algorithmic}[1]
\Require {$i$, $k$, $OV$}
\Ensure { a decision if $N_i$ is a subcomponent union }
\For {$j \gets 1$ to $k$ $\wedge$ $j \neq i$}
\State Find the same variables between $OV_i$ and $OV_j$ 
\Statex \: \quad  and denote it as $X_s$
\If{$\lvert X_s \rvert$ $>$ 0 }
\State $X_r$ $\gets$ $OV_i$ / $X_s$ 
\If{existing $x_i$ in $X_s$ do not interact with $X_r$ }
\State\Return $\emph{true}$
\EndIf
\If{existing $x_i$ in $X_r$ do not interact with $X_s$ }
\State\Return $\emph{true}$
\EndIf
\EndIf
\EndFor 
\State\Return $\emph{false}$
\end{algorithmic}
\end{algorithm}

\begin{algorithm}[!h]
\small
\captionsetup{labelfont={color=black}}
\caption{SD (Subcomponent Detection)}
\label{algorithm_SD}
\begin{algorithmic}[1]
\Require {$i$, $OV$, $N$, $H$, $V$, $\emph{\textbf{x}}_{l,l}$, $\emph{{f}}_{l,l}$, $\emph{\textbf{ub}}$, $\emph{\textbf{lb}}$}
\Ensure { $N$, $OV$ }
\State $X_d$ $\gets$ same variable in $OV_i$ and $H$
\If{$\lvert$$X_d$$\rvert$ $=$ 0}
\State $X_d$ $\gets$ $N_i$ / $OV_i$
\EndIf
\State $X_1$ $\gets$ $N_i$
\While{$\lvert$$X_1$$\rvert$ $=$ $\lvert$$N_i$$\rvert$}
\State $x_d$ $\gets$ Pick a variable from $X_d$ at random
\State $X_1$ $\gets$ INTERACT ($x_d$, $N_i$, $\emph{\textbf{x}}_{l,l}$, $\emph{{f}}_{l,l}$, $\emph{\textbf{ub}}$, $\emph{\textbf{lb}}$)
\EndWhile
\State $N$ $\gets$ $N$ $\bigcup$ $X_1$
\State $X_2$ $\gets$ $N_i$ / $X_1$
\State $OV^{'}$ $\gets$ INTERACT-OV ($X_1$, $X_2$, $\emph{\textbf{x}}_{l,l}$, $\emph{{f}}_{l,l}$, $\emph{\textbf{ub}}$, $\emph{\textbf{lb}}$)
\State $N_i$ $\gets$ $N_i$ / $X_1$ $\bigcup$ $OV^{'}$
\State $OV_i$ $\gets$ INTERACT-OV ($N_i$, $V$, $\emph{\textbf{x}}_{l,l}$, $\emph{{f}}_{l,l}$, $\emph{\textbf{ub}}$, $\emph{\textbf{lb}}$)
\State $OV_1$ $\gets$ INTERACT-OV ($X_1$, $V$, $\emph{\textbf{x}}_{l,l}$, $\emph{{f}}_{l,l}$, $\emph{\textbf{ub}}$, $\emph{\textbf{lb}}$)
\State $OV$ $\gets$ $OV$ $\bigcup$ $OV_1$
\State\Return $N$, $OV$
\end{algorithmic}
\end{algorithm}

\color{black}

The second stage consists of two steps that refine the grouping results.
The first step is to sequentially inspect each subcomponent formed in the previous phase, and the second step is to further decompose inappropriate grouping results.
We designed two methods called subcomponent union detection (SUD) and subcomponent detection (SD) for this purpose.
The details of SUD and SD are presented in Algorithms \ref{algorithm_SUD}-\ref{algorithm_SD}, respectively.

After the OEDG grouping phase, we obtain two key outputs: the set of subcomponents and the set of overlapping variable groups.
These grouping results contain correct groupings as well as incorrect groupings that are the union of multiple subcomponents.
Figure \ref{fig:refine}(a) illustrates a correct grouping result with the subcomponent $\{x_3, x_5, x_6, x_{7}\}$ and shared variables on both sides, $x_3$ and $x_6$ respectively.
Since these shared variables belong to the same subcomponent, they interact with each other.

However, in the case of Fig. \ref{fig:refine}(b), the shared variables on the subcomponent formed by the grouping phase are $x_3$ and $x_9$.
This is an incorrect decomposition result as the two shared variables come from different subcomponents, and there is no interaction between $x_3$ and $x_9$.
Therefore, by detecting interactions between different groups of shared variables for the subcomponent formed in the grouping stage, we can determine whether the formed subcomponent is a union of several subcomponents.
If two variables from different shared variable groups do not interact, the formed subcomponent needs to be further decomposed to improve the grouping accuracy.

After identifying inaccurate grouping results, the next step is to further decompose the union of multiple subgroups.
To achieve this, we identify the non-shared variables present in these merged groups (which may be shared in other groups) and use them as detected variables.
In line topology overlapping problems, we can select the shared variables between the merged group and its adjacent variable groups as the detected variables.
In the example in Fig. \ref{fig:refine}(b), OEDG can select $x_3$ and $x_9$ as the detected variables and find the variables that interact with them in the merged group to form a new subcomponent.
In complex topology problems, where there may be multiple overlapping subcomponents, selecting detected variables can be challenging.
The information regarding the overlapping variable set $OV$ obtained during the problem decomposition phase can guide the selection of appropriate variables.
Variables that exist in $OV_i$ and appear twice in $OV$ are a good choice.
This is because these variables are present in just two subcomponents, and using them as detected variables prevents the identification of an excessive number of interacting variables.
They can be used to identify each subgroup in the merged subcomponent.

The SUD method sequentially inspects each subcomponent formed in the grouping phase.
Firstly, SUD identifies variables in the shared variable group $OV_i$ that overlap with other subcomponents $OV_j$, and we define this group of variables as $X_s$ (Line 2).
The remaining variables in $OV_i$ are defined as $X_r$ (Line 4).
If there are two non-interactive variables in $X_r$ and $X_s$, it indicates that the grouping result of subcomponent $N_i$ is inaccurate, and $N_i$ should be further decomposed.
To ensure detection efficiency, we transform the variable-to-variable detection method into variable-to-set detection to check whether each variable of $X_r$ and $X_s$ interacts with all variables in the other group (Lines 5-10).

After detecting incorrect grouping results, SD proceeds to decompose the subcomponent union.
We construct a promising variable group, denoted as $X_d$, to guide the decomposition of subcomponents (Lines 1-4). 
If there are common variables in $OV_i$ and $H$, then the promising variable group $X_d$ is selected as the intersection of $OV_i$ and $H$.
Otherwise, we utilize the non-shared variable in the merged subcomponent as $X_d$.
SD randomly selects the detected variable $x_d$ from $X_d$.
Subsequently, SD identifies the variables from $N_i$ that interact with $x_d$, forming a new subcomponent denoted as $X_1$  (Lines 6-9).
The newly formed subcomponent is then added to $N$ (Line 10).
SD finds all the shared variables, denoted as $OV^{'}$, between $X_1$ and $X_2$ (Lines 11-12).
SD adds these shared variables $OV^{'}$ to $N_i$ to update the result of the merged subcomponent after separating out the new subcomponent (Line 13).
Finally, the information of the shared variable set $OV$ is updated (Lines 14-16).

After correcting all erroneous grouping results output by the first stage, SD outputs the correct grouping results of subcomponents and shared variables.
In the Section S.III of the Supplementary Material, we provide an example of the OEDG method, particularly highlighting its second stage, in addressing complex topology overlapping problems.

\subsection{Time Complexity Analysis}
In this subsection, we analyze the time complexity in terms of the required fitness evaluations for the OEDG method to decompose an overlapping problem.
OEDG employs the same grouping operation on various topology types overlapping problems. 
For intuitive understanding, we will analyze the line topology overlapping problems as a representative.

Assuming an $n$-dimension line topology overlapping problem comprises $k$ subcomponents, each with $m$ variables, including $m_s$ shared variables and $m_{ns}$ non-shared variables.
In the first problem grouping stage, the function INTERACT is executed fewer than ${2 m \times \log _{2}(n)}$ times to form each subcomponent.
To identify the shared variables in each subcomponent, the function INTERACT-OV is executed less than ${2 k m_{s} \times \log _{2}(m)}$ times.

In the second stage, the SUD method performs variable-to-set interaction detection $2 k m_{s}$ times.
Furthermore, when the SUD method detects merged subcomponents, the SD method decomposes the subcomponents each time with computational complexity akin to the first stage.

Thus, the time complexity of OEDG in terms of the number of FEs is $\mathcal{O}(nlogn)$ on various topology types overlapping problems due to performing variable-to-set and set-to-set interaction detection.
The time complexity of OEDG is equivalent to that of RDG3 and ORDG.
Compared to DG2, the time complexity of OEDG is significantly reduced.

\section{New Overlapping Benchmark Problems}\label{section Benchmark}

The CEC 2013 LSGO \cite{CEC2013} benchmark introduced two novel overlapping functions for the first time.
DOV \cite{DOV} and RDG3 \cite{RDG3} further extended the overlapping size in these overlapping problems.
CBCCO \cite{CBCCO} and DCCMAES \cite{DCCMAES} replaced the basis function and set nonuniform and uniform subgroup sizes.
However, the existing benchmark functions have limitations in terms of topology and separability type, as they all feature a single line topology and possess only additive separability.
To address this shortcoming and introduce more diverse types of overlapping problems, we designed a series of new benchmark functions.

In each benchmark function, all variables are shuffled according to a random sequence $P$, and the shifted vector $\boldsymbol{x}$ is utilized to alter the position of the optimal value for each variable \cite{CEC2010, CEC}.
After the permutation and shift operation, all variables are divided into corresponding subcomponents according to the subgroup size settings.
Each separable subgroup is then transformed into a group of non-separable variables using an orthogonal matrix that undergoes a coordinate rotation operation \cite{CEC2010, CEC}.

\subsection{Topology Type}
The current overlapping benchmarks establish connections between multiple subcomponents exclusively through line topology relationships.
We introduce two novel types of topological structures for overlapping problems: the ring topology and the complex topology.

\subsubsection{Line Topology}
\begin{figure}[hbt!]
        \centering
        \captionsetup{labelfont={color=black}}
        \subfigure[A line topology overlapping problem with uniform subgroup sizes.]{
            \begin{minipage}[b]{0.5\textwidth}
            \centering
            \includegraphics[width=0.9\textwidth]{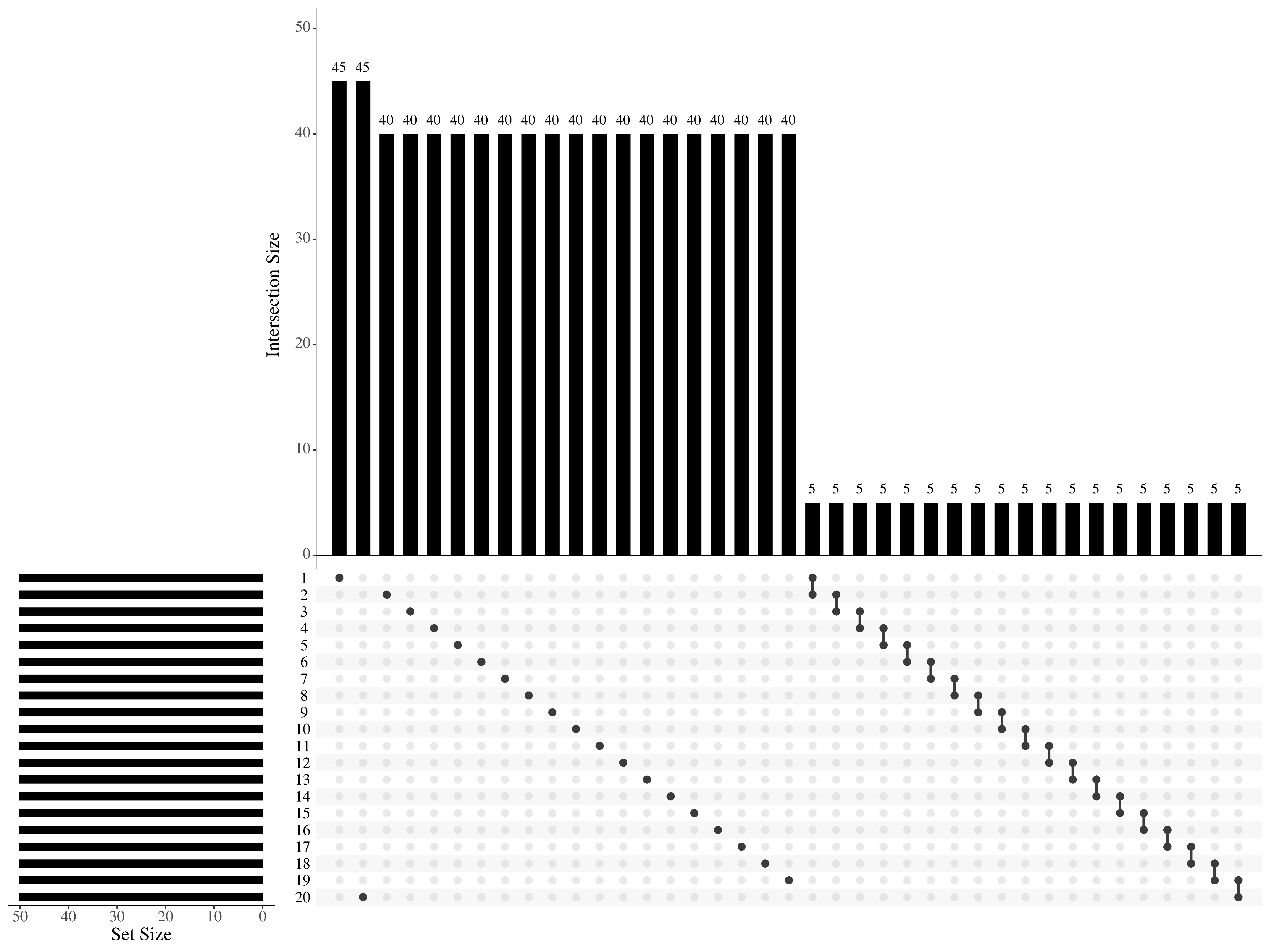}
            \end{minipage}
            }
            \hfill
        \subfigure[A line topology overlapping problem with nonuniform subgroup sizes.]{
            \begin{minipage}[b]{0.5\textwidth}
            \centering
            \includegraphics[width=0.9\textwidth]{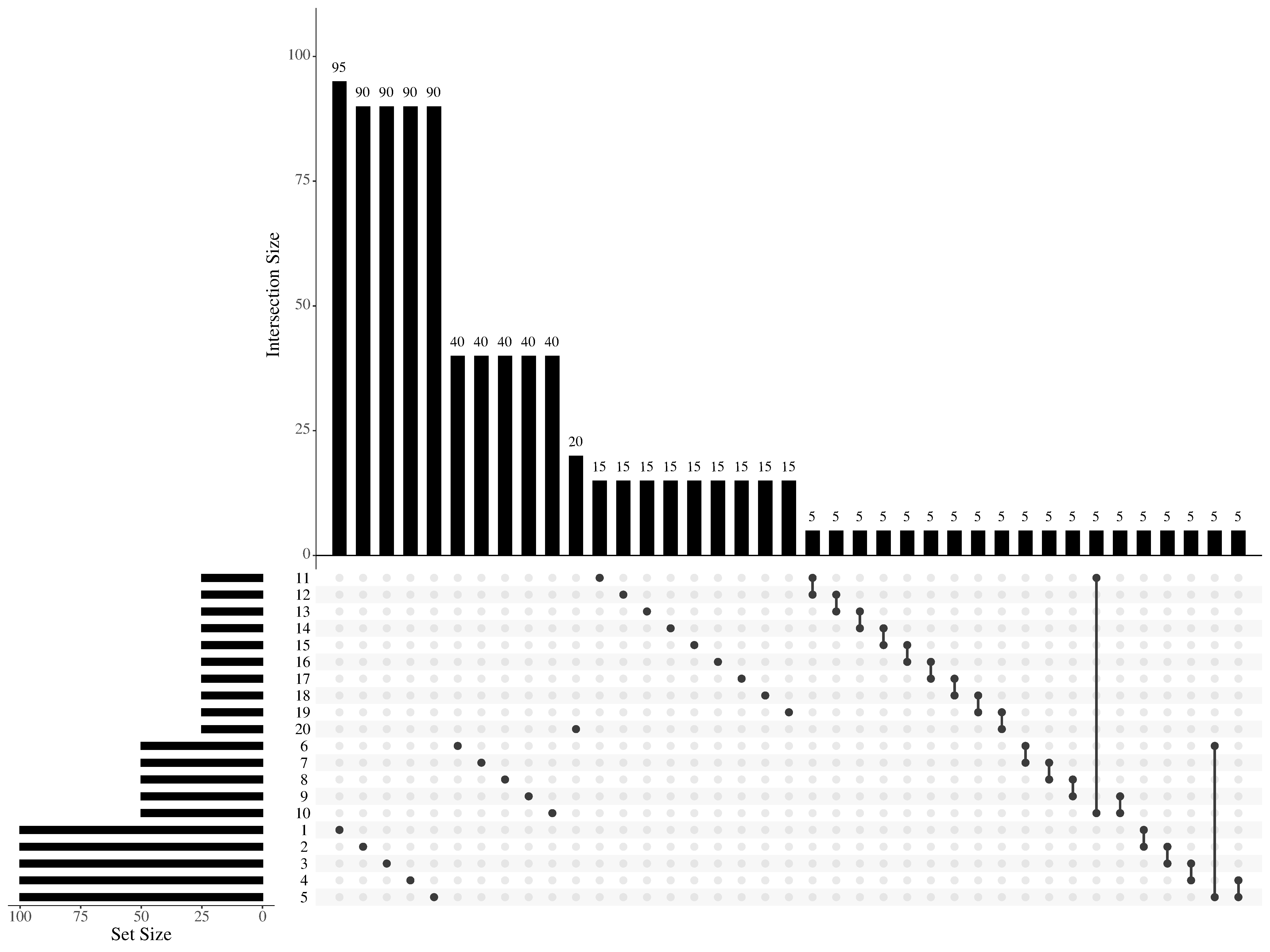}
            \end{minipage}
            }
        \caption{The UpSet diagrams of the line overlapping problems.}
        \label{fig:line}
        \vspace{-0.25cm}
\end{figure}

The benchmark in \cite{DCCMAES} serves as the basis for the line topology overlapping (LTO) benchmark.
Table \ref{tab:benchmark1} illustrates that this benchmark uses three basis functions.
The Schwefel and Elliptic functions are unimodal, while the Rastrigin function is multimodal. 
This benchmark addresses not only uniform and non-uniform group sizes but also conflicting and conforming overlapping problems.
Each problem consists of 20 subcomponents with 5 shared variables between adjacent subcomponents.
Hence, the problems in this benchmark are 905-dimensional.
UpSet diagrams, commonly used in biology to illustrate overlapping relationships between genomes \cite{UPSET}, are employed here to depict the overlapping relationships of subcomponents in the overlapping problems. 
The UpSet diagrams of a line overlapping problem with nonuniform and uniform subgroup sizes are shown in Fig. \ref{fig:line}.

\subsubsection{Ring Topology}

The ring topology overlapping problems share similarities with the line topology overlapping problems.
In fact, the ring topology overlapping (RTO) problem can be constructed by connecting the first and last subcomponents of a line topology overlapping problem.

Therefore, based on the line topology overlapping problem in Table \ref{tab:benchmark1}, we design the corresponding ring topology overlapping problems.
Each problem consists of 900 variables, with 20 subcomponents and 20 5-dimensional shared variable groups.
Due to the page limit, the UpSet diagrams of the ring overlapping problems are shown in Fig. S.5 of the Supplementary Material.


\begin{table}[htbp]
  \centering
  \renewcommand{\arraystretch}{1.25} 
  \captionsetup{labelfont={color=black}} 
  \caption{Large-scale overlapping benchmark functions of line topology and ring topology. The 12 ring topology overlapping (RTO) problems are constructed by connecting the first and last two subcomponents of each line topology overlapping (LTO) problem.  }
  \setlength{\tabcolsep}{4mm}
  \resizebox{1\linewidth}{!}{
    \begin{tabular}{cccc}
    \hline
    Func. & Basic Function  & Character   & Group Size \\
    \hline
    $f_{1}$   & \multirow{4}[2]{*}{Schwefel}  & Conforming & 100×5+50×5+25×10   \\
    $f_{2}$  &  & Conflicting & 100×5+50×5+25×10          \\
    $f_{3}$  &  & Conforming & 50×20          \\
    $f_{4}$  &  & Conflicting & 50×20         \\
    \hline
    $f_{5}$   & \multirow{4}[2]{*}{Elliptic}  & Conforming & 100×5+50×5+25×10 \\
    $f_{6}$   & & Conflicting & 100×5+50×5+25×10         \\
    $f_{7}$   & & Conforming & 50×20        \\
    $f_{8}$    & & Conflicting & 50×20         \\
    \hline
    $f_{9}$  & \multirow{4}[2]{*}{Rastrigin}  & Conforming & 100×5+50×5+25×10   \\
    $f_{10}$  &  & Conflicting & 100×5+50×5+25×10         \\
    $f_{11}$  &  & Conforming & 50×20       \\
    $f_{12}$  &  & Conflicting & 50×20         \\
    \hline
    \end{tabular}}
  \label{tab:benchmark1}%
\end{table}%

\subsubsection{Complex Topology}
In addition to the line and ring topology overlapping problems, there is a more complex category of problems, which we refer to as complex topology overlapping (CTO) problems. 
In such problems, a subcomponent may interact with multiple other subcomponents, and certain variables may simultaneously exist in multiple subcomponents.
We introduce a method to construct scalable complex topology overlapping problems, denoted as CTOC.

\begin{algorithm}[!h]
\small
\captionsetup{labelfont={color=black}}
\caption{CTOC (CTO Construction)}
\label{algorithm_CTOP}
\begin{algorithmic}[1]
\Require { $V$ (all variables), $n$ (number of subcomponents), $s$ (size of subcomponents), $m$ (overlapping size), $p$ (overlapping probability)}
\Ensure { $G$ (set of subcomponents) }
\State $G \gets \{\}$
\State $g_1 \gets $ Pick $s$ variables in $V$  /*Construct the first subcomponent*/
\State $G \gets G \bigcup g_1$
\For {$i \gets 2$ to $n$}
\State $g_k \gets$ Pick a subcomponent from $G$ at random
\State $v \gets$ Pick $m$ variables from $g_k$ at random
\For{$j \gets 1$ to length ($G$) $\wedge$ $j \neq k$}
\If {rand(0,1) $<$ $p$ }
\State $v_j \gets$ Pick $m$ variables from $g_j$ at random
\State $v \gets v \bigcup v_j$
\EndIf
\EndFor
\State $g_i \gets$ Pick variables from $V$ randomly to merge with $v$ 
\State $G \gets G \bigcup g_i$
\EndFor 
\State\Return $G$
\end{algorithmic}
\end{algorithm}

The pseudocode of CTOC is presented in Algorithm \ref{algorithm_CTOP}.
CTOC begins by creating the first subcomponent, denoted as $g_1$, with the $s$ variables in $V$ (Lines 1-3).
For subsequent subcomponents $g_i$, CTOC first chooses one subcomponent $g_k$ from $G$ to establish a connection with (Line 5).
CTOC then randomly selects $m$ variables from $g_k$ to form the shared variable group (Line 6).
To construct more complex overlapping problems, in Lines 7-12, CTOC determines, with a probability $p$, whether $g_i$ will establish a connection with other subcomponents within the set $G$.
The interaction relationship is established by extracting shared variables from the respective subcomponents and incorporating them into $v$ to construct $g_i$ (Line 10).
Finally, the new subcomponent $g_i$ is formed by utilizing the shared variable group $v$ and selecting some variables from $V$ (Line 13).

\begin{figure}[htbp]
\captionsetup{labelfont={color=black}}
\centerline{\includegraphics[width=0.9\linewidth]{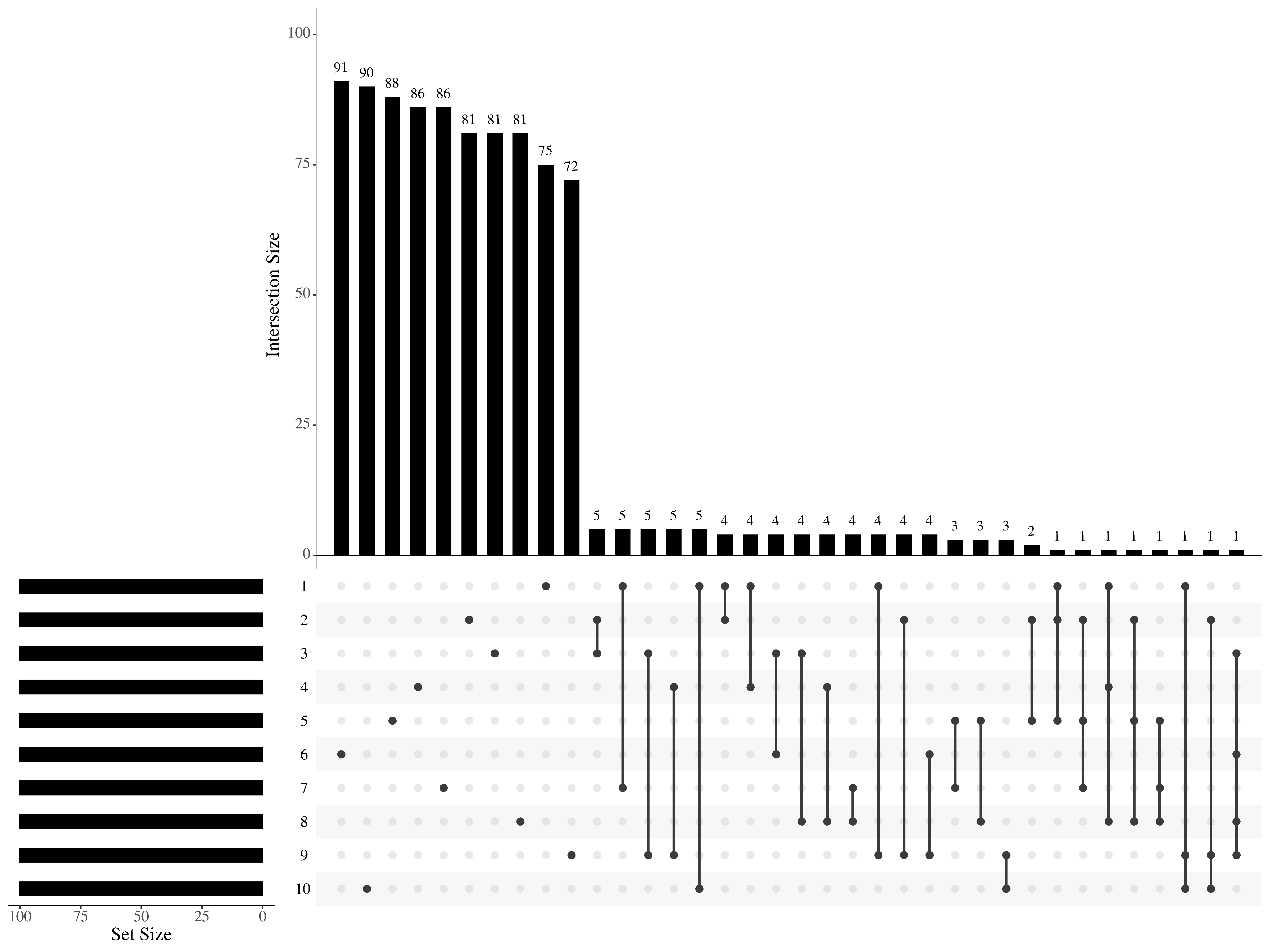}}
\caption{An example of a complex topology overlapping problem generated by CTOC with a parameter setting of $m=5$ and $p=0.2$.}
\label{fig:complex}
\vspace{-0.0cm}
\end{figure}

In CTOC, the overlapping degree of the problems can be adjusted by changing the values of $m$ and $p$. 
The larger $m$ and $p$ are, the higher the overlapping degree is.
Therefore,  this benchmark's design is scalable in terms of the overlapping degree.
CTOC has the capability to produce various types of complex topology.
Figure \ref{fig:complex} shows a complex topology overlapping problem generated by CTOC.

The detailed CTO benchmark utilized in this article is provided in Section S.V of the Supplementary Material.

\subsection{Multi-degree Overlapping Benchmark}
The overlapping degree of the overlapping problems reflects the closeness of the relationship among the subcomponents.
We design a multi-degree overlapping (MDO) benchmark by configuring the overlapping size between subcomponents \cite{CEC2013}.

The MDO dataset contains 15 overlapping problems, as shown in Section S.VII of the Supplementary Material.
$f_1$-$f_{10}$ are line topology and ring topology overlapping problems, each establishing different overlapping degrees by varying the overlapping size.
$f_{11}$-$f_{15}$ are complex topology overlapping problems with varying overlapping degree.
These problems are generated randomly by adjusting the overlapping size $m$ in CTOC.
Moreover, the overlapping degree can be further extended by adjusting the overlapping probability $p$ in CTOC.

\subsection{Non-additively Overlapping Benchmark}
Based on the four non-additively partially separable functions $f_9$-$f_{12}$ in the BNS benchmark \cite{GSG}, we designed eight non-additively overlapping problems which are denoted as NAO.

Following the overlapping problem setting method in CEC2013 \cite{CEC2013}, the original non-additively partially separable problems in BNS are transformed into an overlapping problem using line topology.
Each problem consists of 20 subcomponents with 50 variables, 5 of which overlap between adjacent subcomponents.
In the non-additively overlapping benchmark, $f_1$-$f_{4}$ are the conforming overlapping problems, while $f_5$-$f_{8}$ are the conflicting overlapping problems.
Table \ref{tab:benchmark2} shows the non-additively overlapping benchmark.

\begin{table}[htbp]
  \centering
  \renewcommand{\arraystretch}{1.5} 
  \captionsetup{labelfont={color=black}} 
  \caption{Large-scale non-additively overlapping (NAO) benchmark.}
  \setlength{\tabcolsep}{10pt}
  \resizebox{1\linewidth}{!}{
    \begin{tabular}{p{3.625em}p{6.69em}>{\centering\arraybackslash}p{14.875em}}
    \hline
    Func. & Character  & Function \\
    \hline
    $f_{1}$    & \multirow{4}[2]{*}{Conforming}  & $\prod_{i=1}^{n}F_{\rm{rot\_}\rm{prodsqu}}[\textbf{\emph{z}}_i]-1$ \\
    $f_{2}$    &         &$\prod_{i=1}^{n}F_{\rm{rot\_}prodras}[\textbf{\emph{z}}_i]-1$  \\
    $f_{3}$    &        &$D^2 \cdot \ln \{\sum_{i=1}^{n}F_{\rm{rot\_}\rm{abs}}[\textbf{\emph{z}}_i]+1\}$  \\
    $f_{4}$    &       &$D^2 \cdot \sqrt{ \sum_{i=1}^{n}F_{\rm{rot\_}\rm{elliptic}}[\textbf{\emph{z}}_i]}$  \\
    \hline
    $f_{5}$    & \multirow{4}[2]{*}{Conflicting}   & $\prod_{i=1}^{n}F_{\rm{rot\_}\rm{prodsqu}}[\textbf{\emph{z}}_i]-1$\\
    $f_{6}$    &         &$\prod_{i=1}^{n}F_{\rm{rot\_}prodras}[\textbf{\emph{z}}_i]-1$  \\
    $f_{7}$    &        &$D^2 \cdot \ln \{\sum_{i=1}^{n}F_{\rm{rot\_}\rm{abs}}[\textbf{\emph{z}}_i]+1\}$  \\
    $f_{8}$    &        &$D^2 \cdot \sqrt{ \sum_{i=1}^{n}F_{\rm{rot\_}\rm{elliptic}}[\textbf{\emph{z}}_i]}$  \\
    \hline
    \end{tabular}}
  \label{tab:benchmark2}%
\end{table}%

\section{Experimental Studies}\label{section Experimental}

In this section, we utilize multiple overlapping benchmark problems designed in Section \ref{section Benchmark} to conduct grouping and optimization experiments.
Grouping experiments are carried out to demonstrate the efficiency and accuracy of OEDG in handling large-scale overlapping problems.
Subsequently, we integrate the grouping results generated by OEDG into CBCCO \cite{CBCCO} for optimization. 
Specifically, shared variables are allocated to the subcomponent with the greatest contribution. 
This approach divides the overlapping problem into multiple independent sub-problems. 
Furthermore, we utilized the CC framework to iteratively optimize each subcomponent, leveraging the context vector to facilitate the cooperation and co-evolution among all subcomponents.
Finally, the optimization performance of OEDG is assessed through comparative experiments.

\subsection{Comparison on Decomposition}

In this part, we compare OEDG with state-of-the-art grouping methods (RDG3 \cite{RDG3}, ORDG \cite{ORDG}, DG2 \cite{DG2}) for large-scale overlapping problems in terms of problem decomposition.

ORDG, RDG3, and OEDG utilize an identical approach to establish thresholds for determining separability, following the rules specified in RDG2 \cite{RDG2}.
DG2 employs an automatic setting for the threshold parameter.
The threshold of subcomponent size in RDG3 is set to 50, as suggested in \cite{RDG3}.
The results of the grouping are presented in Tables \ref{tab grouping1}-\ref{tab grouping3} (the results of DG2 are denoted by CBCCO in the tables).

\subsubsection{Accuracy Metric}
The evaluation metrics include the accuracy of the decomposition and the utilization of computational resources.

For an overlapping problem, the $m$ subcomponents are denoted as $\textbf{\emph{subcom}}^*=\{\textbf{\emph{g}}_i^*,...,\textbf{\emph{g}}_m^*\}$, and both shared and non-shared variables are contained in each subcomponent.
Regarding the grouping outcomes, the formed $k$ subcomponents are referred to as $\textbf{\emph{subcom}}=\{\textbf{\emph{g}}_i,...,\textbf{\emph{g}}_k\}$.
The decomposition accuracy of the overlapping problems (DA) is defined as:
\begin{eqnarray}\label{equ NA}
\begin{split}
\textrm{DA}=
\frac{\sum_i^m \max\{|\textbf{\emph{g}}_i^*\cap \textbf{\emph{g}}_1|,|\textbf{\emph{g}}_i^*\cap \textbf{\emph{g}}_2|,..,|\textbf{\emph{g}}_i^*\cap \textbf{\emph{g}}_k|\}}{\sum_i^m |\textbf{\emph{g}}_i^*|}
\end{split}
\end{eqnarray}

The number of function evaluations (FEs) reflects the efficiency of decomposition.
To test the stability of the decomposition method, 30 independent runs were conducted for each algorithm to gather grouping information.
The final evaluation metrics are obtained by averaging the results of 30 independent runs.

\subsubsection{Decomposition on Various Topology Benchmarks} 

\begin{table*}[t]
  \centering
    \captionsetup{labelfont={color=black}} 
  \caption{The decomposition results of each algorithm on the line topology overlapping problems, ring topology overlapping problems, and complex topology overlapping problems. DA represents the decomposition accuracy of the overlapping problems, while FEs represents the number of function evaluations.}
  \resizebox{\linewidth}{!}{
       \renewcommand{\arraystretch}{01}
 \renewcommand{\tabcolsep}{18pt}
\begin{tabular}{cc|cc|cc|cc|cc}
\hline
\multirow{2}[4]{*}{Benchmark} & \multirow{2}[4]{*}{Problem} & \multicolumn{2}{c|}{OEDG} & \multicolumn{2}{c|}{RDG3} & \multicolumn{2}{c|}{ORDG} & \multicolumn{2}{c}{DG2} \bigstrut\\
\cline{3-10}& &DA &FEs &DA &FEs  &DA &FEs &DA &FEs\bigstrut\\
\hline
\multirow{12}[2]{*}{LTO} & $f_{1}$ & \textbf{100\%} & \textbf{23251} & 76.28\% & 16424  & 87.47\% & 15822  & 100\% & 409966  \bigstrut[t]\\
& $f_{2}$ & \textbf{100\%} & \textbf{23089}  & 76.32\% & 16437   & 83.17\%  & 15815 & 100\% & 409966  \\
& $f_{3}$  & \textbf{100\%} & \textbf{24787}  & 70.07\% & 18195  & 80.83\%  & 17729 & 100\% & 409966 \\
& $f_{4}$  & \textbf{100\%} & \textbf{24739}  & 70.48\% & 18295 & 71.50\%  & 17691 & 100\% & 409966 \\
& $f_{5}$   & \textbf{100\%} & \textbf{23287}  & 76.65\% & 16461 & 83.67\%  & 16189  & 100\% & 409966 \\
& $f_{6}$   & \textbf{100\%} & \textbf{22645}  & 76.18\% & 16325  & 83.92\%  & 15822  & 100\% & 409966 \\
& $f_{7}$     & \textbf{100\%} &\textbf{24235}& 69.87\% & 18169  & 74.67\%  & 17681  & 100\% & 409966 \\
& $f_{8}$   & \textbf{100\%} & \textbf{25177}  & 69.68\% & 18128  & 75.83\%  & 18178  & 100\% & 409966 \\
& $f_{9}$ & \textbf{100\%} & \textbf{23005} & 75.55\% & 18198  & 85.67\%  & 16239 & 100\% & 409966 \\
& $f_{10}$    & \textbf{100\%} & \textbf{22882}  & 78.13\% & 16328  & 86.75\%  & 16161  & 100\% & 409966 \\
& $f_{11}$   & \textbf{100\%} & \textbf{24625}  & 70.93\% & 18338  & 77.50\%  & 18202  & 100\% & 409966 \\
& $f_{12}$    & \textbf{100\%} & \textbf{24838}  & 70.73\% & 18195  & 80.00\%  & 18094 & 100\% & 409966 \\
\hline
\multirow{12}[2]{*}{RTO} & $f_{13}$  & \textbf{100\%} & \textbf{23056}  & 74.20\% & 15822  & 71.25\% & 18022  & 100\% & 405451 \bigstrut[t]\\
& $f_{14}$     &\textbf{100\%}& \textbf{22651}  & 74.50\% & 15815  & 71.84\% & 17773  & 100\% & 405451 \\
& $f_{15}$     &\textbf{100\%} & \textbf{24658}  & 67.07\% & 17729  & 54.83\% & 19100  & 100\% & 405451 \\
& $f_{16}$     & \textbf{100\%} & \textbf{24517} & 67.73\% & 17691 & 54.50\% & 18383  & 100\% & 405451 \\
& $f_{17}$     & \textbf{100\%} & \textbf{23041} & 74.70\% & 16188  & 70.08\% & 17705  & 100\% & 405451 \\
& $f_{18}$     &\textbf{100\%}& \textbf{23215}   & 76.57\% & 15822  & 71.00\% & 17891  & 100\% & 405451 \\
& $f_{19}$    &\textbf{100\%} &\textbf{24769}   & 68.77\% & 17680  & 54.67\% & 18549 & 100\% & 405451 \\
& $f_{20}$     & \textbf{100\%} & \textbf{24958} & 68.27\% & 18178 & 54.67\% & 18775  & 100\% & 405451 \\
& $f_{21}$     & \textbf{100\%} & \textbf{22768} & 76.56\% & 16239 & 71.00\% & 17837  & 100\% & 405451 \\
& $f_{22}$    & \textbf{100\%} & \textbf{23170}  & 74.56\% & 16161  & 71.25\% & 18213  & 100\% & 405451 \\
& $f_{23}$    & \textbf{100\%} & \textbf{25006} & 69.42\% & 18202  & 54.33\% & 18812  & 100\% & 405451 \\
& $f_{24}$    & \textbf{100\%} & \textbf{24880}  & 66.68\% & 18094 & 54.67\% & 19168  & 100\% & 405451 \\
\hline
\multirow{12}[2]{*}{CTO} & $f_{25}$ & \textbf{100\%} & \textbf{12389}  & 88.02\% & 14431  & 31.00\% & 8247 & 100\% & 419071 \bigstrut[t]\\
&$f_{26}$    & \textbf{100\%} & \textbf{10066}  & 81.62\% & 14645  & 35.00\% & 8251  & 100\% & 419071 \\
& $f_{27}$    & \textbf{100\%} & \textbf{11215}  & 90.74\% & 14836 & 36.33\% & 8326  & 100\% & 437581 \\
& $f_{28}$    & \textbf{99.30\%} & \textbf{12502}  & 88.33\% & 14399  & 29.97\% & 8244  & 100\% & 414506 \\
& $f_{29}$    & \textbf{100\%} & \textbf{10885} & 83.69\% & 13881  & 30.00\% & 8071  & 100\% & 400961 \\
& $f_{30}$     & \textbf{100\%} & \textbf{12272}  & 88.46\% & 14521  & 35.33\% & 8301  & 100\% & 423661 \\
& $f_{31}$    & \textbf{100\%} & \textbf{13811}  & 85.25\% & 14056 & 31.67\% & 8071  & 100\% & 405451 \\
& $f_{32}$     & \textbf{99.67\%} & \textbf{12117}  & 89.28\% & 14547 & 31.33\% & 8281  & 100\% & 423661 \\ 
& $f_{33}$  & \textbf{100\%} & \textbf{11266} & 90.00\% & 14699  & 40.33\% & 8310  & 100\% & 432916 \\
& $f_{34}$    & \textbf{99.53\%} & \textbf{10189} & 87.81\% & 14388 & 31.00\% & 8448 & 100\% & 415417 \\
& $f_{35}$   & \textbf{100\%} & \textbf{11710} & 90.78\% & 14796 & 47.00\% & 8386  & 100\% & 437581 \\
& $f_{36}$   & \textbf{100\%} & \textbf{14030}  & 86.76\% & 14171 & 30.00\% & 8021  & 100\% & 405451 \\
\hline
\end{tabular}%
    }
  \label{tab grouping1}%
  \vspace{-0.15cm}
\end{table*}%

Table \ref{tab grouping1} shows the decomposition results on various topology benchmarks.
From the results, it can be found that OEDG efficiently and accurately decomposes various topology overlapping problems when compared to three other state-of-the-art grouping methods across all 36 test problems.
This also demonstrates the capability of OEDG's two-stage cooperative decomposition method to address overlapping problems with diverse topology structures. 

Compared to RDG3 and ORDG, OEDG utilizes comparable computational resources but achieves a higher level of decomposition accuracy on the 36 test problems.
CBCCO can accurately identify the structure of overlapping problems and achieves 100$\%$ grouping.
However, its computational resource consumption is excessive, exceeding $4 \times 10^5$ on all problems, which may have some negative impact on optimization.

It is also noticed that OEDG did not achieve complete 100\% decomposition on $f_{25}$, $f_{32}$, and $f_{34}$. 
This is due to the complex structure of such problems, which adversely affects the accuracy of both SUD and SD.

\subsubsection{Decomposition on Multi-degree LSOP} 
The grouping results for overlapping problems with different overlapping degrees are presented in Table \ref{tab grouping2}.
According to the results, it can be found that OEDG achieves the best decomposition performance compared to other methods.

However, OEDG fails to decompose function $f_{14}$ and $f_{15}$ accurately due to the merging of certain subcomponents.
When the number of shared variables increases, the structure of such problems becomes more complex, and the decomposition accuracy of OEDG decreases.
RDG3 and ORDG are also less accurate when decomposing problems with a higher overlapping degree.
Problems with high overlapping degrees pose serious challenges for decomposition. 

DG2 achieves the most accurate results based on its variable-to-variable interaction detection. 
As the overlapping degree increases, the problem size decreases, and the computational resource consumption of DG2 also decreases accordingly.
However, compared with OEDG, DG2 has extremely high computational complexity.

\subsubsection{Decomposition on Non-additively LSOP} 
DG series methods can only decompose additively separable problems and cannot handle non-additively separable problems.
The GSG \cite{GSG} method proposed by us, based on the minimum point shift principle, achieves good performance in decomposing non-additively separable problems.
We replace the interaction detection method in the OEDG framework with GSG, and this modified approach is referred to as OGSG.
The grouping results of non-additively overlapping problems are listed in Table \ref{tab grouping3}.

The experimental results show that OGSG achieves 100\% decomposition accuracy on all problems.
This reflects the potential of the algorithmic framework we designed for handling non-additively overlapping problems.
All other DG-based methods group all variables into a single non-separable group, which cannot accurately identify the subcomponents in the overlapping problems.
This is because DG series methods cannot decompose non-additively separable problems.

\begin{table*}[t]
  \centering
      \captionsetup{labelfont={color=black}} 
  \caption{The decomposition results of each algorithm on the multi-degree overlapping problems. DA represents the decomposition accuracy of the overlapping problems, while FEs represents the number of function evaluations.}
  \resizebox{\linewidth}{!}{
       \renewcommand{\arraystretch}{01}
 \renewcommand{\tabcolsep}{18pt}
\begin{tabular}{cc|cc|cc|cc|cc}
\hline
\multirow{2}[4]{*}{Benchmark} & \multirow{2}[4]{*}{Problem} & \multicolumn{2}{c|}{OEDG} & \multicolumn{2}{c|}{RDG3} & \multicolumn{2}{c|}{ORDG} & \multicolumn{2}{c}{DG2} \bigstrut\\
\cline{3-10}&   &DA   &FEs     &DA   &FEs      &DA   &FEs    &DA   &FEs\bigstrut\\
\hline
\multirow{15}[2]{*}{MDO} & $f_{1}$  & \textbf{100\%} & \textbf{22411} & 76.66\% & 20252  & 77.83\% & 19132 & 100\% & 481672  \bigstrut[t]\\
& $f_{2}$ & \textbf{100\%} & \textbf{23089} & 79.86\% & 17338 & 84.58\%  & 17951  & 100\% & 445097  \\
& $f_{3}$  & \textbf{100\%} & \textbf{25513} & 69.90\% & 18293 & 76.67\%  & 18903  & 100\% & 409966 \\
& $f_{4}$   & \textbf{100\%} & \textbf{25858}  & 71.28\% & 13743  & 89.42\%  & 17306  & 100\% & 328456 \\
& $f_{5}$   & \textbf{100\%} & \textbf{28429}  & 55.68\% & 13038 & 77.00\%  & 17188 & 100\% & 255971 \\
& $f_{6}$   & \textbf{100\%} & \textbf{21877}  & 78.33\% & 18001  & 69.82\%  & 16656  & 100\% & 480691 \\
& $f_{7}$  & \textbf{100\%} &\textbf{24328} & 69.89\% & 19236  & 54.00\%  & 17035  & 100\% & 442271 \\
& $f_{8}$ & \textbf{100\%} &\textbf{25708} & 67.37\% & 18076  & 54.67\%  & 17017  & 100\% & 405451 \\
& $f_{9}$   & \textbf{100\%} & \textbf{25387}  & 70.67\% & 13663  & 70.08\%  & 15659 & 100\% & 320401 \\
& $f_{10}$ & \textbf{100\%} & \textbf{29011}  & 52.68\% & 12648 & 52.93\%  & 15380  & 100\% & 245351 \\
& $f_{11}$  & \textbf{100\%} & \textbf{8806}  & 97.97\% & 15861  & 35.33\%  & 6525 & 100\% & 483637 \\
& $f_{12}$  & \textbf{100\%} & \textbf{11414}  & 95.44\% & 15518  & 50.00\%  & 6832  & 100\% & 470936 \\
& $f_{13}$  & \textbf{100\%} & \textbf{11274} & 88.59\% & 14542  & 30.67\%  & 6755  & 100\% & 428276 \\
& $f_{14}$  & \textbf{98.77\%} & \textbf{19547} & 76.50\% & 12521 & 27.67\%  & 6327 & 100\% & 337432 \\
& $f_{15}$  & \textbf{94.33\%} & \textbf{23871} & 59.54\% & 10189  & 23.67\%  & 5582  & 100\% & 255971 \\
\hline
\end{tabular}%
    }
  \label{tab grouping2}%
    \vspace{-0.25cm}
\end{table*}%

\begin{table*}[t]
  \centering
      \captionsetup{labelfont={color=black}} 
  \caption{The decomposition results of each algorithm on the non-additively overlapping problems. DA represents the decomposition accuracy of the overlapping problems, while FEs represents the number of function evaluations.}
  \resizebox{\linewidth}{!}{
       \renewcommand{\arraystretch}{1}
 \renewcommand{\tabcolsep}{12pt}
\begin{tabular}{cc|cc|cc|cc|cc|cc}
\hline
\multirow{2}[4]{*}{Benchmark} & \multirow{2}[4]{*}{Problem} & \multicolumn{2}{c|}{OGSG} & \multicolumn{2}{c|}{OEDG} & \multicolumn{2}{c|}{RDG3} & \multicolumn{2}{c|}{ORDG} & \multicolumn{2}{c}{DG2} \bigstrut\\
\cline{3-12}      &     &DA   &FEs        &DA   &FEs     &DA   &FEs      &DA   &FEs     &DA   &FEs \bigstrut\\
\hline
\multirow{8}[2]{*}{NAO} & $f_{1}$  & \textbf{100\%} &\textbf{62871}  & 5.00\% & 5431 & 5.00\% & 5422  & 5.00\% & 5422 & 5.00\% & 409966  \bigstrut[t]\\
& $f_{2}$   & \textbf{100\%} & \textbf{61217} & 5.00\% & 5431  & 5.00\% & 5422  & 5.00\% & 5422  & 5.00\% & 409966  \\
& $f_{3}$   & \textbf{100\%} & \textbf{62354} & 5.00\% & 5431   & 5.00\% & 5422 & 5.00\% & 5422  & 5.00\% & 409966 \\
& $f_{4}$  & \textbf{100\%} & \textbf{63024} & 5.00\% & 5431  & 5.00\% & 5422  & 5.00\% & 5422 & 5.00\% & 409966 \\
& $f_{5}$   & \textbf{100\%} & \textbf{62372} & 5.00\% & 5431  & 5.00\% & 5422 & 5.00\% & 5422  & 5.00\% & 409966 \\
& $f_{6}$   & \textbf{100\%} & \textbf{61223}  & 5.00\% & 5431  & 5.00\% & 5422 & 5.00\% & 5422  & 5.00\% & 409966 \\
& $f_{7}$   & \textbf{100\%} &\textbf{62768} & 5.00\% & 5431 & 5.00\% & 5422  & 5.00\% & 5422  & 5.00\% & 409966 \\
& $f_{8}$   & \textbf{100\%} &\textbf{65144} & 5.00\% & 5431  & 5.00\% & 5422  & 5.00\% & 5422 & 5.00\% & 409966 \\
\hline
\end{tabular}%
    }
  \label{tab grouping3}%
\end{table*}%

\color{black}

\subsection{Comparison on Optimization}

In the optimization experiments, we select six state-of-the-art algorithms for comparison. 
These include three non-decomposition-based methods: CSO \cite{CSO}, GL-SHADE \cite{GLSHADE}, and MLSHADE-SPA \cite{SHADE}. 
CSO employs a competition mechanism to enhance population diversity and convergence. 
Both the GLSHADE and MLSHADE-SPA algorithms incorporate a range of techniques to bolster their performance on LSGO problems. 
GLSHADE demonstrated superior results on previous overlapping benchmark problems, while MLSHADE-SPA secured the second place in the CEC 2018 LSGO competition. 
Additionally, we considered three decomposition-based CC series methods—RDG3 \cite{RDG3}, ORDG \cite{ORDG}, and CBCCO \cite{CBCCO}—specifically tailored for addressing large-scale overlapping problems.

In the original articles, the covariance matrix adaptation evolution strategy (CMA-ES) \cite{CMAES} was utilized as the optimizer for RDG3 and CBCCO \cite{RDG3, CBCCO}.
To ensure fairness, we also employed CMA-ES as the optimizer for ORDG, using the more effective CBCCO framework instead of its original CC framework.

For our proposed OEDG, we likewise employed CBCCO as the CC framework for the optimization stage. 
This choice was made because OEDG can accurately explore the structure of the overlapping problems by identifying all the subcomponents and overlapping variables, as shown in Section III. 
This makes it particularly well-suited for the CBCCO framework.

The optimization termination condition is set to $3 \times 10^6$ FEs. 
The optimization results are obtained through 30 independent runs.
Wilcoxon's rank-sum test is conducted on the optimization results at a significance level of 0.05. 
Results with a significant advantage are highlighted in bold.

\subsubsection{Optimization on Various Topology Benchmarks}

\begin{table*}[htbp]
  \centering
  \captionsetup{labelfont={color=black}}
  \caption{Comparative statistical optimization results between OEDG and corresponding algorithms on LTO, RTO, and CTO benchmarks. The symbols ``W'', ``T'', and ``L'' indicate that OEDG is significantly better than, statistically similar to, and significantly worse than the compared algorithm, respectively.}
   \resizebox{\linewidth}{!}{
  \setlength{\tabcolsep}{15pt}
    \begin{tabular}{cccccccc}
    \hline
    Statistical term & OEDG  & CSO   & GL-SHADE & MLSHADE-SPA & RDG3  & ORDG  & CBCCO \bigstrut\\
    \hline
    W/T/L & -    & 27/5/4  & 27/4/5  & 29/3/4  & 25/2/9  & 26/1/9  & 17/19/0 \bigstrut\\
    \hline
    \end{tabular}%
    }
  \label{tab:comp}%
    \vspace{-0.35cm}
\end{table*}%

Due to the page limit, the detailed optimization results for various topology benchmarks (LTP, RTO, and CTO) are presented in Section S.VI of the Supplementary Material.
The overall comparative statistical results are listed in Table \ref{tab:comp}. 
The results indicate that OEDG can efficiently and accurately identify the structure of overlapping problems and seamlessly be integrated into the extended CBCCO framework, resulting in its superior optimization results across all 36 problems.

For line and ring topology overlapping problems, OEDG achieves the best optimization results based on the accurate identification of the problem structure and integration of a contribution-based variable grouping and optimization framework.
However, for complex topology overlapping problems, despite the higher decomposition accuracy of the OEDG method, its optimization results are unsatisfactory.
The reason is that OEDG encounters challenges in establishing rational and efficient shared variable allocation schemes for complex topology overlapping problems.
Moreover, the contribution-based CC framework utilized by OEDG is not well-suited for addressing such problems.
This optimization framework may destabilize the optimization process, leading to significant fluctuations in the outcomes achieved by OEDG.

For multimodal problems $f_9$-$f_{13}$,  CSO outperforms other methods due to its excellent exploration capabilities.
For $f_{26}$, $f_{28}$, and $f_{30}$-$f_{32}$, RDG3 achieves the best optimization results, the reason is that it sets the threshold to randomly assign shared variables to adjacent subcomponents, effectively controlling subcomponent sizes.
Additionally, its optimization framework is simple and efficient.
For $f_{33}$-$f_{36}$, ORDG decomposes complex topology overlapping problems into subcomponents with a large size.
This prevents the incorrect assignment of shared variables, resulting in the best optimization results.

\subsubsection{Optimization on Multi-degree LSOP} 

\begin{table*}[htbp]
  \centering
  \captionsetup{labelfont={color=black}}
  \caption{Comparative statistical optimization results between OEDG and corresponding algorithms on MDO benchmark. The symbols ``W'', ``T'', and ``L'' indicate that OEDG is significantly better than, statistically similar to, and significantly worse than the compared algorithm, respectively.}
   \resizebox{\linewidth}{!}{
  \setlength{\tabcolsep}{15pt}
    \begin{tabular}{cccccccc}
    \hline
    Statistical term & OEDG  & CSO   & GL-SHADE & MLSHADE-SPA & RDG3  & ORDG  & CBCCO \bigstrut\\
    \hline
    W/T/L & -    & 10/1/4  & 9/0/6  & 10/1/4  & 12/0/3  & 12/0/3  & 8/7/0 \bigstrut\\
    \hline
    \end{tabular}%
    }
  \label{tab:MDOCOM}
\end{table*}%

Due to the page limit, the detailed optimization results for the multi-degree overlapping problems are shown in Section S.VII of the Supplementary Material.
The overall comparative statistical results are presented in Table \ref{tab:MDOCOM}. 

Overall, OEDG outperforms the algorithms in comparison in terms of optimization results.
It efficiently identifies the structure of overlapping problems, allocates shared variables, and optimizes subcomponents based on contributions, leading to improved optimization results.

However, OEDG performs less effectively when dealing with complex topology overlapping problems with a higher overlapping degree and overlapping problems containing smaller subcomponents.
These problems pose significant challenges for the shared variable allocation mechanism and contribution-based CC framework in the OEDG optimization process.

\subsubsection{Optimization on Non-additively LSOP}

\begin{table*}[htbp]
  \centering
  \captionsetup{labelfont={color=black}}
  \caption{Comparative statistical optimization results between OGSG and corresponding algorithms on NAO benchmark. The symbols ``W'', ``T'', and ``L'' indicate that OGSG is significantly better than, statistically similar to, and significantly worse than the compared algorithm, respectively.}
   \resizebox{\linewidth}{!}{
  \setlength{\tabcolsep}{25pt}
    \begin{tabular}{cccccc}
    \hline
    Statistical term & OGSG  & CSO   & GL-SHADE & MLSHADE-SPA & DG-SERIES \bigstrut\\
    \hline
    W/T/L & -    & 4/2/2  & 6/1/1  & 3/4/1  & 5/2/1 \bigstrut\\
    \hline
    \end{tabular}%
    }
  \label{tab:NAOCOM}%
    \vspace{-0.35cm}
\end{table*}%

Due to the page limit, the experimental results for non-additively overlapping problems are provided in Section S.VIII of the Supplementary Material.
The overall comparative statistical results are shown in Table \ref{tab:NAOCOM}. 
All DG-based decomposition methods group the eight non-additively overlapping problems into a single group.
Therefore, we use DG-SERIES to represent the optimization results of DG-series methods (OEDG, RDG3, ORDG, DG2) in Table S.IV.
The grouping results of the OGSG method on the NAO benchmark are also integrated into CBCCO for optimization. 

OGSG achieved superior optimization results compared to other algorithms.
Specifically, $f_1$ and $f_5$ are overlapping problems constructed by partially multiplicatively separable functions and all methods demonstrated good optimization performance on these two problems.
However, the optimization effect of OGSG may be insufficient for certain problems due to the absence of subgroup weights in the NAO benchmark. 
This absence directly affects the shared variable allocation mechanism of OEDG, thereby diminishing its optimization performance.

\subsubsection{Exploratory Optimization Experiment} 

To further investigate the impact of shared variables on overlapping problems and explore when breaking the connection between subcomponents is more beneficial for optimization, we conducted a series of optimization experiments between OEDG and non-decomposition-based methods on overlapping problems with a series of higher overlapping degrees.
We extend the overlapping degree in the MDO benchmark.
Additionally, we integrate accurate subcomponent information into the OEDG framework for optimization and compare the obtained results with non-decomposition-based methods.
The detailed benchmarks and optimization results can be found in Section S.IX of the Supplementary Material.

In the line topology overlapping problems, OEDG yields superior optimization results when the overlapping size is below 15.  
However, beyond an overlapping size of 20, non-decomposition-based methods outperform OEDG. 
This is attributed to the increased overlapping degree leading to a reduction in problem dimensionality.
Non-decomposition-based methods excel at handling low-dimensional problems, making decomposition unnecessary.
In addition, higher overlapping degrees erode the distinct overlapping property, bringing them closer to fully non-separable problems. 
In such cases, rational selection of optimization subcomponents becomes challenging.

When confronted with complex topology overlapping problems with a series of higher overlapping degrees, OEDG performs less effectively than the three non-decomposition methods. 
This can be attributed to the CBCCO optimization framework employed by OEDG, which struggles to address such problems.
In contrast, GL-SHADE combines global search and local search techniques, making it perform better on unimodal problems ($f_1$-$f_{10}$). 
CSO has demonstrated exceptional performance on multimodal problems ($f_{11}$-$f_{20}$), thanks to its outstanding exploration capabilities.

Therefore, based on the above discussion, we should consider whether to break the linkage between shared variables from two perspectives: the topology type and overlapping degree of the overlapping problems.
Moreover, when dealing with overlapping problems characterized by a complex topology and a high overlapping degree, exploring a more rational problem decomposition method and utilizing a hybrid optimization framework could be a promising avenue to address these challenges.

\color{black}

\section{Conclusion}\label{section Conclusion}

In this article, we propose an enhanced differential grouping method called OEDG for large-scale overlapping problems. 
This article presents a comprehensive study on large-scale overlapping problems compared to state-of-the-art studies. 
We consider various properties of large-scale overlapping problems, including the topology structure, overlapping degree, separability, etc. 

OEDG consists of two stages: the problem grouping stage and the grouping refinement stage.
The first stage ensures the efficiency of OEDG, while the second stage guarantees its accuracy.
During the grouping stage, a variable is randomly sampled, and the variables directly interacting with it are identified to form a subcomponent.
Subsequently, the overlapping variables within this subcomponent are identified, and the relevant information is recorded. 
In the grouping refinement stage, the information obtained in the first stage is utilized to identify subcomponents that are not fully decomposed due to random sampling, using the SUD method.
These subcomponents are further decomposed utilizing the SD method to obtain more precise results.

We create a series of novel overlapping benchmark problems, designed based on various topology types, varying overlapping degrees, and extended separability.
These benchmarks address the limitations of overlapping benchmark functions in previous work.
We have verified the performance of the proposed OEDG algorithm through a series of experiments on the benchmarks we designed.
The experimental results demonstrate that OEDG achieves high accuracy in grouping while consuming fewer computational resources. 
Furthermore, we have validated the optimization performance of OEDG by integrating it into the CBCCO framework.
Due to its accurate exploration of the problem structure and low consumption of computational resources, OEDG significantly outperforms the compared methods in terms of optimization results.

In the future, we will endeavor to design effective CC frameworks to optimize large-scale overlapping problems.
This may involve applying more rational computational resource allocation strategies, integrating mechanisms for shared variable characterization into CC, and dynamically allocating shared variables.
Additionally, addressing the challenges of conflicting overlapping problems and complex topology overlapping problems represents a new research direction.

\footnotesize

\bibliography{manuscript_references}

\end{document}